\def\year{2022}\relax
\documentclass[letterpaper]{article} 
\usepackage{aaai22}  
\usepackage{times}  
\usepackage{helvet}  
\usepackage{courier}  
\usepackage[hyphens]{url}  
\usepackage{graphicx} 
\usepackage{natbib}  
\usepackage{caption} 
\DeclareCaptionStyle{ruled}{labelfont=normalfont,labelsep=colon,strut=off} 
\usepackage{algorithm}
\usepackage{algorithmic}

\usepackage{newfloat}
\usepackage{listings}
\floatstyle{ruled}
\newfloat{listing}{tb}{lst}{}
\floatname{listing}{Listing}
\title{Effective and Interpretable Information Aggregation with Capacity Networks}

\author {
     Markus Zopf
}
 \affiliations{
     NEC Laboratories Europe\\
     Heidelberg, Germany \\
     markus.zopf@neclab.eu
 }

\usepackage{booktabs}
\usepackage{subcaption}
\usepackage{tikz}
\usepackage{tikzscale}
\usepackage{pgfplots, pgfplotstable}
\usepgfplotslibrary{statistics}
\usepackage{float}
\usepackage{makecell}
\usepackage{ifthen}
\usepackage{mathtools}
\usepackage{multirow}
\usepackage{bbm}
\usepackage{tikz}
\usetikzlibrary{shapes}
\tikzset{mark options=solid}
\usepackage{colortbl}
\usepackage{enumitem}
\usepackage{wrapfig}
\usepackage{amsfonts}

\newcommand{\mnistred}{Unique Sum}
\newcommand{\mnistsyn}{Weighted Triangular}
\newcommand{\mnistredsyn}{Unique Sum + Synergy Bonus}

\newcommand{\fmnisttri}{Triangular Count}
\newcommand{\fmnistuniqecount}{Unique Count}

\newcommand{\sentiment}{Sentiment Analysis}

\newcommand{\mnistredshort}{US}
\newcommand{\mnistsynshort}{WTri}
\newcommand{\mnistredsynshort}{US+S}
\newcommand{\mnistmulshort}{Mult}

\newcommand{\fmnisttrishort}{TriC}
\newcommand{\fmnistuniqecountshort}{UC}

\newcommand{\sentimentshort}{Sent}

\newcommand{\encoder}{\mathsf{encoder}}
\newcommand{\decoder}{\mathsf{decoder}}

\newcommand{\dset}{\mbox{DeepSet}}
\newcommand{\att}{\mbox{Attention}}

\newcommand{\smallst}{\mbox{Set Trans.}}
\newcommand{\st}{\mbox{Set Trans. L}}

\newcommand{\capacity}{\mathcal{C}} 

\newcommand{\rnn}{\mbox{RNN}}
\newcommand{\rnnc}{\mbox{$\capacity$-RNN}}

\newcommand{\lstm}{\mbox{LSTM}}
\newcommand{\lstmc}{\mbox{$\capacity$-LSTM}}

\newcommand{\gru}{\mbox{GRU}}
\newcommand{\gruc}{\mbox{$\capacity$-GRU}}

\newcommand{\objrepfunc}[1][]{\ifthenelse{\equal{#1}{}}{o}{o({#1})}} 
\newcommand{\initobjrep}[1][]{\ifthenelse{\equal{#1}{}}{\varphi}{\varphi({#1})}} 

\newcommand{\instance}{x} 
\newcommand{\instanceutility}{y} 
\newcommand{\bag}{X} 
\newcommand{\bagutility}{Y} 
\newcommand{\groundset}{\mathcal{G}} 

\newcommand{\latsetrep}{Z} 
\newcommand{\latinstancerep}{z} 

\newcommand{\featureagg}[1][]{\ifthenelse{\equal{#1}{}}{f}{f({#1})}} 

\newcommand{\setutilityfunc}[1][]{\ifthenelse{\equal{#1}{}}{v}{v({#1})}} 
\newcommand{\objutilityfunc}[1][]{\ifthenelse{\equal{#1}{}}{u}{u({#1})}} 
\newcommand{\latobjutility}{y} 

\DeclarePairedDelimiter\size{\lvert}{\rvert}
\newcommand{\abs}[1]{\text{abs}(#1)}

\definecolor{nice_red}{rgb}{0.82, 0.1, 0.26}
\definecolor{nice_green}{rgb}{0.0, 0.5, 0.0}
\definecolor{nice_blue}{rgb}{0.0, 0.5, 1.0}

\newcommand{\p}[1]{\textcolor{nice_red}{#1}}
\newcommand{\n}[1]{\textcolor{nice_green}{#1}}

\newcommand{\lightmidrule}{\arrayrulecolor{black!30} \midrule \arrayrulecolor{black}}

\newcommand{\reals}{\mathbb{R}} 
\newcommand{\set}[1]{\{{#1}\}}
\newcommand{\powerset}[1]{\mathcal{P}({#1})}
\newcommand{\coun}{\text{count}}
\newcommand{\indicator}{\mathbbm{1}}

\begin{document}

\maketitle

\begin{abstract}
How to aggregate information from multiple instances is a key question multiple instance learning. Prior neural models implement different variants of the well-known encoder-decoder strategy according to which all input features are encoded a single, high-dimensional embedding which is then decoded to generate an output. In this work, inspired by Choquet capacities, we propose Capacity networks. Unlike encoder-decoders, Capacity networks generate multiple interpretable intermediate results which can be aggregated in a semantically meaningful space to obtain the final output. Our experiments show that implementing this simple inductive bias leads to improvements over different encoder-decoder architectures in a wide range of experiments. Moreover, the interpretable intermediate results make Capacity networks interpretable by design, which allows a semantically meaningful inspection, evaluation, and regularization of the network internals.
\end{abstract}

\section{Introduction}
In many important problems, a label is not given for each individual instance but only a set of instances. For instance, in histopathological classification of breast cancer images \cite{Sudharshana2019}, only one label that indicates if a patient has cancer or not is given for a set of images. Similarly, in drug activity prediction \cite{Dietterich1997}, a label is only provided for a set of molecule shapes, but not for individual molecule shapes. In sentiment analysis, only a single label may be given for a document that is comprised of multiple sentences, without labeling information on the sentence level \cite{Pappas2017,McAuley2012}.

Problems with this characteristic are in the focus of \emph{multiple instance learning} (MIL) \cite{Dietterich1997,Maron1998,Ray2001,Carbonneau2018}. More formally, in MIL, an input consists of a set of instances $\bag \subset \groundset$, where $\groundset$ is a ground set of individual instances. The goal is to learn a set function $f$ from training pairs $(\bag_i, \bagutility_i)$ that maps input sets to set labels. Supervision (i.e. labeling information) for individual instances is not provided. 

\begin{figure*}
	\centering
	\begin{subfigure}{0.5\textwidth}
		\includegraphics[width=\textwidth]{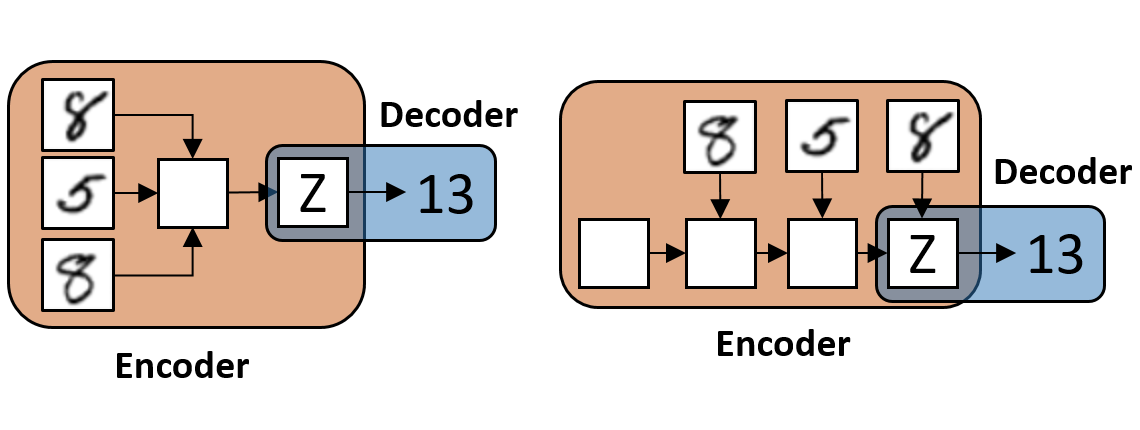}
		\caption{Parallel and sequential encoder-decoder networks}
		\label{fig:encoder_decoder_illustration}
	\end{subfigure}
	\hspace{1.5cm}
	\begin{subfigure}{0.3\textwidth}
		\includegraphics[width=\textwidth]{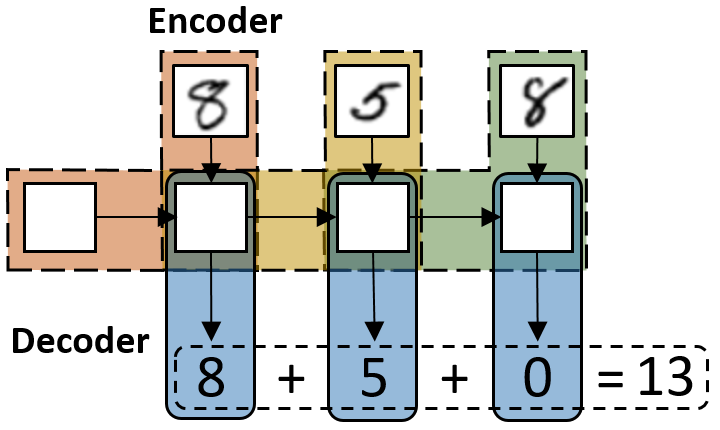}
		\caption{Capacity networks}
		\label{fig:capacity_network_illustration}
	\end{subfigure}
	\caption{High-level illustration of parallel and sequential encoder-decoders (\ref{fig:encoder_decoder_illustration}) and the newly proposed Capacity networks (\ref{fig:capacity_network_illustration}) applied to an instance of a popular MNIST-based multiple instance regression problem called 'unique sum' \cite{Murphy2019,Kalra2020}. The input consists of a set of MNIST images (in this example the three images \includegraphics[scale=0.35]{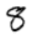}, \includegraphics[scale=0.40]{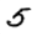}, and \includegraphics[scale=0.35]{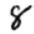}) and the output equals to the sum of uniquely appearing MNIST images. Both encoder-decoder architectures generate a set embedding $\latsetrep$ which is decoded to obtain the output and, hence, serves as interface between encoder and decoder. In contrast, Capacity networks apply a decoder three times to generate three intermediate results, each of which modeling the added value of instance $\instance_i$ with respect to all already observed instances $\instance_1, \dots, \instance_{i-1}$. We use red, yellow, and green to illustrate that the encoder network is applied three times.}
	\label{fig:encoder_decoder_capacity_networks}
\end{figure*}

A key question in MIL is how to aggregate the information from the multiple instances in the input sets. Prior neural approaches for MIL such as \citet{Zaheer2017,Ilse2018,Murphy2019,Lee2019} implement different variants of the encoder-decoder strategy. Encoder-decoders use an encoder networks to aggregate the features of all individual instances $\instance_1, \dots, \instance_n \in \bag$ into a single, high-dimensional set embedding $\latsetrep$. The set embedding is then decoded by a decoder network to generate an output $\bagutility$. We provide a high-level illustration of sequential and parallel encoder-decoder networks in Figure~\ref{fig:encoder_decoder_illustration}.

In this work, we present an alternative approach to aggregate information from multiple instances and demonstrate its effectiveness and interpretability. Inspired by a sequential decomposition of Choquet capacities \cite{Choquet1954,Sugeno1974}, we propose to decompose multiple instance learning tasks with sets of size $n$ into $n$ smaller sub-problems (i.e. one for each instance in the input) and to produce a meaningful intermediate result for each sub-problem. More specifically, for each instance $i$, an neural module generates an intermediate set embedding which is immediately decoded by another neural module to generate an intermediate result. Each intermediate result models the added value of instance $\instance_i$ with respect to all already seen instances $\instance_1, \dots, \instance_{i-1}$. Like in Choquet capacities, all intermediate results can be simply summed to obtain the final output. In Figure~\ref{fig:encoder_decoder_capacity_networks}, we illustrate the architectures of the resulting family of neural networks, which we denote as Capacity networks, and show how it differs from encoder-decoder architectures.

Capacity networks have two major advantages over their encoder-decoder counterparts: \emph{improved effectiveness} and \emph{interpretability by design}. More specifically, we find that Capacity networks achieve better results than prior works in a wide range of experiments. A potential explanation for the improved performance is that Capacity networks are better able to leverage the compositional nature of MIL problems. Moreover, Capacity networks are interpretable by design. Prior encoder-decoder architectures are end-to-end black boxes. In contrast, Capacity network generate multiple meaningful intermediate results that can easily be interpreted. The improved interpretability allows to evaluate if the trained models are right for the right reasons. Moreover, a quantitative evaluation of the intermediate results is possible. Furthermore, the intermediate results allow to inject semantically meaningful prior knowledge into the training process via regularization terms. To summarize, our contributions are as follows:

\begin{enumerate}
	\item We propose Capacity networks that implement an alternative information aggregation strategy for multiple instance learning. Our key idea is to generate an interpretable, scalar-valued intermediate result for each instance that represents its added value with respect to all already observed instances and to sum all generated intermediate results to obtain the final prediction.
	\item We show that Capacity networks are more effective than prior encoder-decoder architectures at learning challenging set functions. Moreover, we show that Capacity networks perform better for large sets, for varying set sizes, with smaller amounts of training data, and in a real-world sentiment analysis dataset.
	\item We demonstrate the interpretability of Capacity networks, which enable a quantitative evaluation of the network internals and the implementation of a semantically meaningful regularization.
\end{enumerate}

\section{Capacity Networks}
\label{sec:capacity_networks}
In the following, we discuss an iterative decomposition of Choquet capacities, which motivate the fundamental architecture of Capacity networks.

\subsection{Iterative Decomposition of Choquet Capacities}
\label{sec:choquet_integration}
Let $\groundset=\set{x_1, \dots, x_m}$ be a set of instances. A set function $\mu : \powerset{\groundset} \rightarrow [0, \infty)$ is called \emph{(Choquet) capacity} \cite{Choquet1954,Sugeno1974} if $\mu(\emptyset) = 0$ and $\mu(A) \leq \mu(B)$ for $A \subseteq B \subseteq \groundset$. Unlike measures such as the Lebesgue measure, capacities are not additive, which is essential to model non-additive aggregation effects, i.e. problems where $\mu(A \cup B) = \mu(A) + \mu(B), \forall A,B \in \groundset$ with $A \cap B = \emptyset$ does not hold in general. A capacity $\mu$ is called \emph{superadditive} if $\mu(A \cup B) \geq \mu(A) + \mu(B)$ and \emph{subadditive} if $\mu(A \cup B) \leq \mu(A) + \mu(B)$, for $A \cap B = \emptyset, A, B \in \powerset{\groundset}$.

Given a capacity $\mu$, it is natural to ask how large the individual contribution of a single instance $x_i$ is to a set $X=\set{x_1, \dots, x_n}, x_i \notin X$. More formally: Given $X \subset \groundset, x_i \notin X$, how large is $\mu(X \cup \set{x_i}) - \mu(X)$? In the additive case (like in measures), the difference always equals to $\mu(\set{x_i})$. However, in the non-additive case, the added value of $x_i$ to $X$ depends on $X$. The motivation to identify the individual contribution of an instance to a set utility naturally leads to the following sequential decomposition:

\begin{equation}
	\label{eq:incremental_capacities}
	\begin{split}
		\mu(X) = & \;\;\;\;\, [\mu(\emptyset \cup \set{x_1}) - \mu(\emptyset)] \\
		& + [\mu(\set{x_1} \cup \set{x_2}) - \mu(\set{x_1})] \\
		& + \dots \\
		& + [\mu((X) - \mu(X \setminus \set{x_n})]. \\
	\end{split}
\end{equation}

Each row in Equation~\ref{eq:incremental_capacities} represents the \emph{added value} of a single instance with respect to all previous instances. Equation~\ref{eq:incremental_capacities} can be summarized as 

\begin{equation}
	\label{eq:sum_of_added_values}
	\mu(X) = \sum_{i=1}^{\size{X}} \nu_i, \text{ with } \nu_i = \mu(C_{i-1} \cup \set{x_i}) - \mu(C_{i-1}),
\end{equation}

where $C_{i-1} = \set{x_1, \dots, x_{i-1}}$, and $C_0 = \emptyset$. The first summand $\nu_1 = \mu(C_0 \cup \set{x_1}) - \mu(C_0)$ equals $\mu(\set{x_1}))$. In other words, $\nu_1$ represents the intrinsic value of $x_1$. Similarly, we obtain $\mu(\set{x_1, x_2}) - \mu(\set{x_1})$ for $\nu_2$ which can be interpreted as estimating the added value of $x_2$ with respect to $x_1$. Finally, $\nu_n$ equals to the added value of the last instance $\instance_n$ with respect to all other instances $\instance_1, \dots, \instance_{n-1}$. The decomposition requires an ordering of the instances in $X$. Otherwise, it is unclear which instance should be considered first, second, etc. To this end, any arbitrary total ordering can be used. The input in Figure~\ref{fig:encoder_decoder_capacity_networks}, $X = \set{\includegraphics[scale=0.35]{images/mnist_8_2.png}, \includegraphics[scale=0.40]{images/mnist_5_1.png}, \includegraphics[scale=0.35]{images/mnist_8_1.png}}$), can, for instance, be decomposed as
\begin{equation}
	\label{eq:incremental_capacities_example}
	\begin{split}
		\mu(X) & = \mu(\set{\includegraphics[scale=0.35]{images/mnist_8_2.png}}) \\ & +  [\mu(\set{\includegraphics[scale=0.35]{images/mnist_8_2.png}, \includegraphics[scale=0.40]{images/mnist_5_1.png}}) - \mu(\set{\includegraphics[scale=0.35]{images/mnist_8_2.png}})] \\ & +  [\mu(\set{\includegraphics[scale=0.35]{images/mnist_8_2.png}, \includegraphics[scale=0.40]{images/mnist_5_1.png}, \includegraphics[scale=0.35]{images/mnist_8_1.png}}) - \mu(\set{\includegraphics[scale=0.35]{images/mnist_8_2.png}, \includegraphics[scale=0.40]{images/mnist_5_1.png}})].
	\end{split}
\end{equation}

According to this decomposition and the 'unique sum' task described in Figure~\ref{fig:encoder_decoder_capacity_networks}, we first compute the individual contribution of image $\includegraphics[scale=0.35]{images/mnist_8_2.png}$ (which is $8$), then we compute the contribution of the next image $\includegraphics[scale=0.40]{images/mnist_5_1.png}$ with respect to the already considered image $\includegraphics[scale=0.35]{images/mnist_8_2.png}$ (which is $5$), and finally compute the contribution of last image $\includegraphics[scale=0.35]{images/mnist_8_1.png}$ with respect to $\includegraphics[scale=0.35]{images/mnist_8_2.png}$ and $\includegraphics[scale=0.40]{images/mnist_5_1.png}$ (which is $0$, since class '8' has already appeared once).

\subsection{Network Architecture}
\label{sec:network_architecture}
Inspired by the sequential decomposition of capacities, we now define a new family of neural networks for multiple instance learning. Key idea is to introduce an inductive bias such that neural networks mimic the discussed sequential decomposition of capacities. To this end, we propose to design networks such that they generate an intermediate result $\instanceutility_i$ after reading instance $\instance_i$ which models the added value of instance $\instance_i$ to the set of all already seen instances $\set{\instance_1, \dots, \instance_{i-1}}$. As in Equation~\ref{eq:sum_of_added_values}, the output of the network for a set $\bag$ can be obtained by simply computing the sum of all intermediate results. In terms of encoder and decoder functions, we define Capacity networks as

\begin{align}
	\latinstancerep_i & = \encoder(x_i, \latinstancerep_{i-1}) \label{eq:uan_encoder} \\
	\instanceutility_i & = \abs{\decoder(\latinstancerep_i) \label{eq:uan_decoder}} \\
	\bagutility & = \sum_{i = 1}^{\size{\bag}} \instanceutility_i \label{eq:uan_aggregation},
\end{align}

where $\latinstancerep_{0}$ is an initial state and $\abs{.}$ denotes the absolute value of the decoder output. We compute $\abs{.}$ to model the fact that each $\nu_i \geq 0$, since $\mu(C_{i-1} \cup \set{x_i}) \geq \mu(C_{i-1})$ due to the monotonicity of $\mu$. In principal, this additional inductive bias is not essential for the networks architecture and can be removed if monotonicity should not be enforced. Furthermore, we do not explicitly model subtrahend and minuend explicitly but estimate the difference directly. Again, this is a non-essential design choice that can be modified in future work.

Capacity networks are closely connected the sequential decomposition of capacities presented in Equation~\ref{eq:sum_of_added_values}. Each application of the decoder in Equation~\ref{eq:uan_decoder} corresponds to one $\nu_i$ in Equation~\ref{eq:sum_of_added_values}, which represents the added value of the $i$-th instance with respect to all already observed instances.
%
%
%
%
%
%
Similar to the sequential decomposition of capacities in Equation~\ref{eq:sum_of_added_values}, Capacity networks require a sequential ordering of the inputs. Otherwise, it is not possible to model the added value of an instance with respect to already seen instances. Any sequential network such as RNNs, LSTMs, and GRUs can be used as basis to implement the proposed inductive bias. Hence, we do not present a specific neural networks but rather a new family of neural networks. Capacity networks are not guaranteed to be permutation invariant. This is, however, not necessarily a limitation of the presented idea. Sequential approaches have been demonstrated to show strong performance for MIL in prior works. Furthermore, several methods exist to mitigate or completely remove the permutation sensitivity of permutation sensitive networks as described in the next section. Permutation sensitivity can even be viewed as advantage since our networks can also be used if the output depends on the order of the input instances (e.g. in ordered sets or sequences).

A key distinction between prior encoder-decoder architectures and the newly presented idea is that encoder-decoders do not produce intermediate results, whereas Capacity networks generate $n$ intermediate results, each of which modeling the added value of a specific instance. Another way to describe this distinction is to note that standard encoder-decoders only produce a single latent set embedding which is fed only once into a decoder to produce a scalar-valued output. In contrast, Capacity networks apply a decoder several times (once for each produced latent state $\latinstancerep_i$) to produce many latent scalar-valued intermediate results. As a consequence, encoder-decoders aggregate information from different instances in an uninterpretable high-dimensional feature space, whereas Capacity networks aggregate interpretable scalar-valued intermediate results. The inductive bias to generate scalar-valued intermediate results can also be viewed as introducing multiple information bottlenecks into the network architectures since the Capacity networks need to compress the feature representation to a meaningful, one-dimensional value multiple times.

\section{Related Work}
\label{sec:related_work}
Capacities and Choquet integration have already inspired other works in machine learning before. \citet{Beliakov2008} fit values of the discrete Choquet integral with linear programming techniques. \citet{Tehrani2012a} use the Choquet integral to model monotone nonlinear aggregations for binary classification. \citet{Tehrani2012} use the Choquet integral in a pairwise preference learning scenario. \citet{Dias2018} replace pooling layers in convolutional neural networks with Choquet integration. In contrast to our work, they consider a classification setup and do not consider multiple instance learning. \citet{Karczmarek2018} and \citet{Karczmarek2019} use the Choquet integral to build an ensemble of classifiers for face recognition. Similarly, \citet{Anderson2018} and \citet{Scott2017} build an ensemble of CNNs.

Many works on multiple instance classification \cite{Dietterich1997,Maron1998} and regression \cite{Ray2001,Herbrich2006TrueSkill} make strong task-specific assumption \cite{Carbonneau2018}. For instance, \citet{Dietterich1997} assume that the label of a set is positive if at least one instance in the set is positive (commonly known as the standard MIL assumption \cite{Carbonneau2018}). Similarly, \citet{Ray2001} assume that the target variable depends only on a single primary instance. TrueSkill \cite{Herbrich2006TrueSkill} assumes additivity. These independence assumptions simplify the problem substantially, since models do not have to consider interdependencies between instances. In contrast, we consider more challenging non-additive problems. Prior neural approaches for MIL such as \citet{Zaheer2017,Ilse2018,Murphy2019,Kalra2020} implement different variants of the encoder-decoder strategy. The approaches can be grouped into two groups, depending on whether the encoder network reads all input instances in parallel or sequentially (see Figure~\ref{fig:encoder_decoder_illustration}).

Parallel architectures usually use a permutation invariant pooling operation such as sum, mean, or max to generate a set embedding \cite{Lee2019}. \citet{Zaheer2017} show that permutation invariant architectures can be described in a fairly simple framework consisting of a network $\phi$ that maps instance into a latent feature space, a sum pooling operation, and a decoder network $\rho$. In this case, the encoder can be written as $\encoder(X) = \sum_{x \in X} \phi(x) = \latsetrep$ and the decoder is simply $\decoder(\latsetrep) = \rho(\latsetrep)$. Even though using a sum is sufficiently expressive from a theoretical point of view, \citet{Soelch2019} show that the networks can be highly sensitive to the choice of the aggregation function in practice. Similarly, PointNets \cite{Qi2017} and PointNetST \cite{Segol2020} are universal approximations of invariant and equivariant set functions, respectively. \citet{Soelch2019} propose a trainable recurrent aggregation function based on the read-process-write architecture \cite{Vinyals2016}. Similarly, \citet{Ilse2018} and \citet{Yang2020a} use an attention-based weighted aggregation. The encoder can be written as $\encoder(X) = \sum_{x \in X} a_i \cdot \phi(x_i)$ where $a_i = \frac{\exp{h(\instance_i)}}{\sum_{j=1}^{\size{X}} \exp{h(\instance_j)}}$ and $h$ is an implementation-specific function to transform the input features. \citet{Lee2019} present the self-attention-based \cite{Vaswani2017} Set Transformer and describe how their architecture fits into the encoder-decoder design pattern. Similarly, Deep Message Passing on Sets \cite{Shi2020} can also model interactions between different instances.

Sequential architectures generate the embedding $\latsetrep$ with a sequential encoder network. For instance, \citet{Vinyals2016} compute attention weights based on the hidden state of an LSTM and the input instances. Hence, the encoder can be formulated as $\encoder(X) = \sum a_i x_i$, where $a_i$ are the last layer's attention weights. Since sequential encoder-decoders consume one instance at each timestep $t$ \cite{Cho2014,Luong2015}, the output may change when the input is permuted \cite{Vinyals2016}. Several approaches exist to use permutation-sensitive models as basis for creating permutation-invariant models. A very simple yet effective approach is to order the objects in the input set according to an arbitrary total ordering $\pi$ before they are fed into the permutation-sensitive model \cite{Niepert2016}. \citet{Zhang2020} present a pooling method for sets of feature vectors which can be used to construct permutation-equivariant models. Janossy pooling \cite{Murphy2019} aggregates the outputs of permutation sensitive models to obtain a permutation-invariant model or an approximation thereof. \citet{Meng2019} extend Janossy pooling and use an LSTM to construct a hierarchical feature aggregation network for set-of-sets problems. \citet{Zhang2019} learn how to order the instances in sets before they are fed into an LSTM. Similarly, \citet{Mena2018} learn to reconstruct scrambled objects with Gumbel-Sinkhorn networks. Neural networks with external memories \cite{Das1992,Schmidhuber1992,Schmidhuber1993} such as RNNSearch \cite{Bahdanau2015}, Memory Networks \cite{Weston2015}, and Neural Turing Machines \cite{Graves2014} are alternative permutation sensitive approaches. Similarly, AMRL \cite{Beck2020} is an permutation invariant external memory for reinforcement learning. Unlike the intermediate results in Capacity networks, the external memory is an uninterpretable accumulator of knowledge.

\begin{table}
	\centering
	\begin{tabular}{l @{\hspace{0.2cm}} c @{\hspace{0.2cm}} c @{\hspace{0.2cm}} c @{\hspace{0.2cm}} c @{\hspace{0.2cm}} c @{\hspace{0.2cm}} c}
		& \fmnistuniqecountshort{} & \mnistredshort{} & \mnistsynshort{} & \mnistredsynshort{} & \fmnisttrishort{} & \sentimentshort{} \\
		\toprule
		\rnn	&	0.56	&	0.91	&	1.73	&	6.76	&	4.09	&	1.98 \\
		\lstm	&	0.22	&	0.43	&	0.87	&	2.68	& \textbf{1.01}	&	1.58 \\
		\gru	&	0.21	&	0.47	&	0.86	&	2.44	&	3.02	&	0.74 \\
		\lightmidrule
		\rnnc{} (ours)  &	0.22	&	0.43	&	0.57	&	2.57	&	3.17	&	0.70 \\
		\lstmc{} (ours)	&	0.25	&	0.27	&	0.59	&	\textbf{1.61}	&	1.02	&	\textbf{0.67} \\
		\gruc{} (ours)	&	\textbf{0.18}	&	\textbf{0.25}	&	\textbf{0.48}	&	1.63	&	2.13	&	0.71 \\
		\lightmidrule
		\dset	&	0.51	&	0.27	&	2.90	&	2.19	&	2.03	&	0.79 \\
		\att	&	0.73	&	0.51	&	1.01	&	2.44	&	3.12	&	1.98 \\
		\smallst&	0.30	&	1.91	&	8.72	&	14.74	&	1.22	&	0.92 \\
		\st		&	0.28	&	1.72	&	3.01	&	12.61	&	1.19	&	0.91 \\
		\bottomrule
	\end{tabular}
	\caption{Mean squared errors of three different runs for different architectures in the multiple instance learning datasets. Median results and standard deviation can be found in the supplementary material.}
	\label{tab:full_training_results}
\end{table}

\section{Effectiveness Evaluation}
\label{sec:experiments}
In the following, we evaluate the effectiveness of Capacity networks and compare it with their non-capacity counterparts and other prior architectures. To this end, we implement three Capacity networks based on $\rnn$s, $\lstm$s \cite{Hochreiter1997}, and $\gru$s \cite{Cho2014}. We refer to the resulting Capacity networks as $\rnnc$, $\lstmc$, and $\gruc$, respectively. Capacity networks have the same number of trainable parameters as their non-capacity counterparts. Following \citet{Zaheer2017}, we use 3 fully connected layers for decoder and encoder. In all experiments, we feed the instances in random order into the networks. Additionally, we use a DeepSet \cite{Zaheer2017}, two Set Tranformers \cite{Lee2019} with different sizes\footnote{We use the code provided by the authors \cite{Lee2019} at \url{github.com/juho-lee/set_transformer}.} and an attention-based network \cite{Luong2015} as additional reference models. Mean squared error (MSE) is used as loss and evaluation metric. In all experiments, we report the average of three runs with different random seeds. Adam \cite{Kingma2015} is used as optimizer as it has been used by prior works \cite{Murphy2019} and shows good performance across several tasks and setups \cite{Schmidt2020}. We perform the same hyperparameter optimization for all models. Additional details can be found in the supplementary material.

\subsection{Datasets}
We follow prior works \cite{Zaheer2017,Ilse2018,Murphy2019,Kalra2020} and generate input sets $\bag_i$ and corresponding labels $\bagutility_i$ based on the MNIST \cite{LeCun1998} dataset. Similar to \citet{Murphy2019} and \citet{Kalra2020}, we generate sets with 10 instances and use 100k input sets for training, and 10k sets for validation and test. We generate five challenging datasets:

In the \emph{subadditive} MNIST-based \textbf{\mnistred{} (\mnistredshort)} task \cite{Murphy2019}, the goal is to learn the set function $f_{\text{\mnistredshort{}}} (\bag) = \sum_{c=0}^9 c \cdot \indicator_{\bag}(c)$, where $\indicator_{\bag}(c)$ denotes the indicator function. In principle, the mapping from MNIST classes to indexes $c$ is arbitrary. We map each class to the index corresponding to its numerical value, i.e. class '1' is mapped to index $c=1$, etc. \textbf{\mnistsyn{} (\mnistsynshort)} models \emph{superadditive} effects according to $f_{\text{\mnistsynshort{}}} (\bag) = \sum_{c=0}^9 c \cdot T_{\coun_{\bag}(c)}$, where $\coun_{\bag}(c)$ denotes the number of times MNIST image class $c$ appears in set $\bag$ and $T_m$ is the $m$-th triangular number, i.e. $T_m = \frac{m \cdot (m+1)}{2}$. \textbf{\mnistredsyn{} (\mnistredsynshort)} combines both \emph{subadditive and superadditive} effects in a single dataset by computing the label according to

\begin{equation}
	\label{eq:mnistredsyn}
	f_{\text{\mnistredsynshort{}}} (\bag) = \sum_{c=0}^9 c \cdot \indicator_{\bag}(c) + \sum_{\mathclap{\substack{i=0, j=1\\i<j}}} 10 \cdot \indicator_{\set{c_i, c_j}} \in P, \\
\end{equation}

where $\indicator_{\set{c_i, c_j}} \in P$ indicates whether set $\set{c_i, c_j}$ with MNIST classes $c_i$ and $c_j$ appear in a randomly generated set of pairs $P$.

In addition to tasks based on MNIST, we also generate two datasets based on Fashion-MNIST \cite{Xiao2017}. In \textbf{\fmnistuniqecount{} (\fmnistuniqecountshort{})}, the task is to count the number of uniquely appearing Fashion-MNIST classes. \textbf{\fmnisttri{} (\fmnisttrishort{})} goes beyond simply counting and computes the set labels according to $f_{\text{\fmnisttrishort{}}} (\bag) = \sum_{c=0}^9 T_{\coun_{\bag}(c)}$.

In addition to previously used tasks, we also perform experiments on a real-world \textbf{\sentiment{} (\sentimentshort{})} dataset \cite{Pappas2017}. The dataset is based on work by \citet{McAuley2012} and available online\footnote{Available at \url{https://www.idiap.ch/paper/hatdoc}.}. A set in this dataset represents a document containing multiple sentences. The goal is to predict the overall sentiment of the documents. We use 300-dimensional feature representation to encode each sentence with a standard library\footnote{Available at \url{github.com/UKPLab/sentence-transformers}.} and use the overall sentiment as described in \citet{Pappas2017} as label.

\subsection{Results}
\label{sec:full_training_data}

\begin{table*}
	\centering
	\begin{tabular}{l r r r r r r r r r}
		& \multicolumn{3}{c}{\mnistredshort{}} &  \multicolumn{3}{c}{\mnistsynshort{}} & \multicolumn{3}{c}{\mnistredsynshort{}} \\
		
		\cmidrule(lr){2-4} \cmidrule(lr){5-7} \cmidrule(lr){8-10}
		& 30k & 50k & 70k & 30k & 50k & 70k & 30k & 50k & 70k \\	
		\cmidrule(lr){2-10}
		\rnn			&	1.42	&	1.01	&	1.04	&	11.68	&	5.91	&	3.02	&	63.87	&	8.27	&	7.88\\
		\rnnc{} (ours)	&	\n{-0.45}	&	\n{-0.46}	&	\n{-0.54}	&	\n{-6.21}	&	\n{-3.78}	&	\n{-2.11}	&	\n{-53.98}	&	\n{-3.67}	&	\n{-4.81}\\
		\lightmidrule
		\lstm	&	0.84	&	0.60	&	0.42	&	5.05	&	2.41	&	1.27	&	5.71	&	3.55	&	2.58\\
		
		\lstmc{} (ours)	&	\n{\textbf{-0.14}}	&	\n{\textbf{-0.24}}	&	\n{\textbf{-0.12}}	&	\n{-0.73}	&	\n{-0.80}	&	\n{-0.45}	&	\p{+0.43}	&	\n{\textbf{-0.80}}	&	\n{\textbf{-0.85}}\\
		\lightmidrule
		\gru			&	0.76	&	0.50	&	0.40	&	4.18	&	1.95	&	1.01	&	6.19	&	4.01	&	2.82\\
		\gruc{} (ours)	&	$\pm$0.00	&	\n{-0.07}	&	\p{+0.05}	&	\n{\textbf{-0.45}}	&	\n{\textbf{-0.59}}	&	\n{\textbf{-0.35}}	&	\n{\textbf{-0.86}}	&	\n{-1.12}	&	\n{-0.88}\\
	\end{tabular}
	\caption{MSE of three Capacity networks and their non-capacity counterparts for different training set sizes. Green and red numbers indicate lower and higher relative MSE with respect to the non-Capacity counterparts, respectively. Best results are highlighted in bold face. Absolute values and results for other architectures can be found in the supplementary material.}
	\label{tab:var_training_data}
\end{table*}

We report the results for all datasets and all architectures in Table~\ref{tab:full_training_results} and make several key observations. First, Capacity networks achieve overall the best results. $\lstmc$ and $\gruc$ perform better than $\rnnc$, indicating that Capacity networks benefit from a more complex hidden state update approach. Second, and more importantly, Capacity networks perform in almost all cases better than their directly comparable non-capacity counterparts (i.e. $\rnn$s, $\lstm$s, and $\gru$s). These results provide clear evidence that the introduced inductive bias has a systematic positive effect on the network performance. Third, datasets \mnistredshort{} and \mnistsynshort{} are easier to learn than \mnistredsynshort{} and \fmnisttrishort{}. More difficult tasks (\mnistredsynshort{}) and more difficult images (\fmnisttrishort{}) can be explanations for this. Hence, using these datasets in future works will contribute to a better evaluation of neural MIL works. Furthermore, we find that permutation invariant networks do not have a systematic advantage over permutation sensitive models albeit the set functions to learn are permutation invariant. This is an surprising additional insight that can be investigated further in future work.

\subsection{Experiments with Larger Set Sizes}
\label{sec:larger_set_sizes}

\begin{figure}
	\begin{tikzpicture}
		\begin{axis}[
			width=\linewidth,
			legend pos=north west,
			legend style={font=\tiny,legend columns=2},
			xlabel=Set Size,
			xtick={10,20,30,40},
			ymode=log,
			xlabel near ticks,
			ylabel near ticks]
			
			\addplot[color=nice_blue,dashed,mark=*] coordinates {(10, 1.726) (20, 21.504) (30, 166.297) (40, 632.863)};
			\addplot[color=nice_blue,solid,mark=*] coordinates {(10, 0.573) (20, 5.462) (30, 22.803) (40, 53.555)};
			
			\addplot[color=nice_red,dashed,mark=triangle] coordinates {(10, 0.868) (20, 7.238) (30, 14.28) (40, 27.029)};
			\addplot[color=nice_red,solid,mark=triangle] coordinates {(10, 0.591) (20, 1.597) (30, 4.68) (40, 10.892)};
			
			\addplot[color=nice_green,dashed,mark=square] coordinates {(10, 0.858) (20, 4.713) (30, 12.221) (40, 28.036)};
			\addplot[color=nice_green,solid,mark=square] coordinates {(10, 0.484) (20, 1.513) (30, 3.21) (40, 8.85)};
			
			\addplot[color=black,dotted,mark=o] coordinates {(10, 2.896) (20, 18.641) (30, 51.49) (40, 94.563)};
			\addplot[color=black,dotted,mark=square] coordinates {(10, 1.009) (20, 7.417) (30, 25.014) (40, 49.747)};
			
			\addplot[color=black,mark=o] coordinates {(10, 8.717) (20, 37.139) (30, 95.767) (40, 179.005)};
			\addplot[color=black,mark=square] coordinates {(10, 3.006) (20, 10.117) (30, 26.267) (40, 38.221)};
			
			\legend{\rnn{},\rnnc{},\lstm{},\lstmc{},\gru{},\gruc{},\dset{},\att{}, \smallst{},\st{}}
		\end{axis}
	\end{tikzpicture}
	\caption{Mean squared error for all architectures and different set sizes in the \mnistsynshort{} dataset.}
	\label{fig:larger_set_sizes_MNIST_syn}
\end{figure}
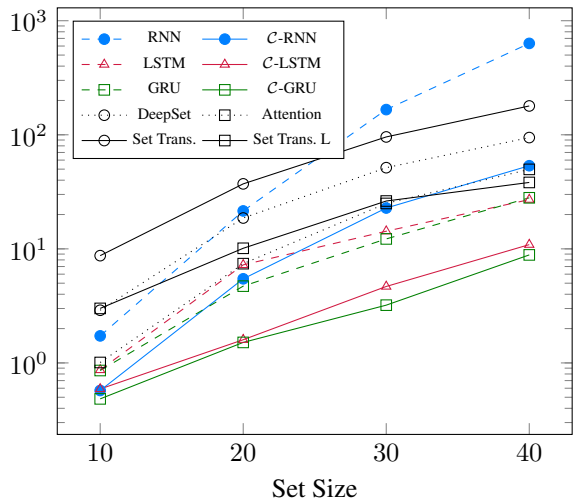

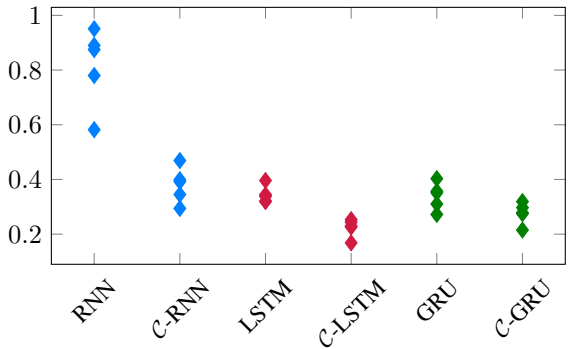
\begin{figure}
	\centering
	\begin{tikzpicture}
		\begin{axis}
			[width=\linewidth,
			height=5cm,
			xtick={1,2,3,4,5,6},
			xticklabels={\rnn{}, \rnnc{}, \lstm{}, \lstmc{}, \gru{}, \gruc{}},
			xticklabel style={font=\small, rotate=45},
			xlabel near ticks,
			ylabel near ticks
			]
			
			\addplot[color=nice_blue,mark=diamond*,mark size=3pt,draw=none] coordinates {(1,0.582)(1,0.780)(1,0.875)(1,0.890)(1,0.951)};
			
			\addplot[color=nice_blue,mark=diamond*,mark size=3pt,draw=none] coordinates {(2,0.294)(2,0.345)(2,0.392)(2,0.399)(2,0.469)};
			
			\addplot[color=nice_red,mark=diamond*,mark size=3pt,draw=none] coordinates {(3,0.32)(3,0.32)(3,0.338)(3,0.344)(3,0.396)};
			
			\addplot[color=nice_red,mark=diamond*,mark size=3pt,draw=none] coordinates {(4,0.168)(4,0.226)(4,0.230)(4,0.244)(4,0.253)};
			
			\addplot[color=nice_green,mark=diamond*,mark size=3pt,draw=none] coordinates {(5,0.272)(5,0.31)(5,0.351)(5,0.357)(5,0.403)};
			
			\addplot[color=nice_green,mark=diamond*,mark size=3pt,draw=none] coordinates {(6,0.215)(6,0.275)(6,0.278)(6,0.297)(6,0.319)};
		\end{axis}
	\end{tikzpicture}
	\caption{Mean squared error for 5 different random permutations of the instances in the \mnistredshort{} dataset.}
	\label{fig:permutation_sensivity}
\end{figure}

In Figure~\ref{fig:larger_set_sizes_MNIST_syn}, we report the performance of all architectures for the \mnistsynshort{} task with larger set sizes. These experiments are more challenging and go beyond the scope of prior works such as \citet{Murphy2019} and \citet{Kalra2020} which only consider sets up to 10 instances. Again, we observe that Capacity networks achieve smaller errors that their non-capacity counterpart (dashed vs. solid lines). It should be noted that we use a log scale for the MSE, which means that the performance improvement when using Capacity networks is not approximately constant, but increases substantially with increasing set sizes. We also report the performance of other prior architectures and can confirm that they usually perform worse for larger set sizes.

\subsection{Training with Smaller Amounts of Training Data}
\label{sec:var_training_data}
In Table~\ref{tab:var_training_data}, we report the performance for different amounts of training data. Similar to the previous experiments, we observe that Capacity networks perform better than their non-capacity counterparts in almost all cases (indicated by green negative numbers). Absolute numbers and results of other models can be found in the supplementary material.

\subsection{Permutation Sensitivity Evaluation}
\label{sec:permutation_sensitivity}
We perform additional experiments to evaluate how sensitive the sequential architectures are with respect to the order of the input instances. In Figure~\ref{fig:permutation_sensivity}, we plot the results of five different runs with randomly permuted instances. The experiments show that the permutation sensitivity of all models is rather low, except for the \rnn{}. In addition to the results in Table~\ref{tab:full_training_results}, the results show that permutation sensitivity seems to be a minor important issue in these datasets. Moreover, we find that Capacity networks can further reduce the permutation sensitivity for all three base architectures.

\section{Interpretability Experiments}
\label{sec:interpretability}
In the following, we demonstrate the improved interpretability of Capacity networks and further  advantages of Capacity networks that are related to the improved interpretability.

\begin{table}
	\centering
	\begin{tabular}{r @{\hspace{0.25cm}} c @{\hspace{0.25cm}} c @{\hspace{0.25cm}} c @{\hspace{0.25cm}} c @{\hspace{0.25cm}} c @{\hspace{0.25cm}} r}
		\multicolumn{7}{c}{ } \\
		Input &
		\scalebox{0.40}{\includegraphics{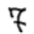}} &
		\scalebox{0.40}{\includegraphics{images/mnist_8_1.png}} &
		\scalebox{0.40}{\includegraphics{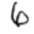}} &
		\scalebox{0.40}{\includegraphics{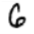}} &
		\scalebox{0.40}{\includegraphics{images/mnist_8_2.png}} & \multicolumn{1}{c}{$\sum$} \\
		
		Expected & 7.00    & 8.00    & 6.00    & 0.00    & 0.00    & 21.00 \\
		\midrule
		\multicolumn{1}{c}{\rnnc{}}  & 6.97 & 7.99 & 6.01 & 0.01 & 0.00 & 20.98\\
		\multicolumn{1}{c}{$\Delta$} & 0.03 & 0.01 & 0.01 & 0.01 & 0.00 & 0.06\\
		\lightmidrule
		\multicolumn{1}{c}{\lstmc{}} & 6.83 & 7.88 & 6.03 & 0.12 & 0.17 & 21.03 \\
		\multicolumn{1}{c}{$\Delta$} & 0.17 & 0.12 & 0.03 & 0.12 & 0.17 & 0.61\\
		\lightmidrule
		\multicolumn{1}{c}{\gruc{}}  & 6.58 & 7.47 & 6.11 & 0.44 & 0.40 & 21.00 \\
		\multicolumn{1}{c}{$\Delta$} & 0.42 & 0.53 & 0.11 & 0.44 & 0.40 & 1.90 \\
	\end{tabular}
	\caption{Example for improved interpretability in the \mnistred{} dataset with 5 instances per set. We show MNIST images (first row), expected intermediate results (second row), predicted intermediate results, and distance to the expected values ($\Delta$ rows).}
	\label{tab:interpretability}
\end{table}

\subsection{Improved Interpretability}
\label{sec:improved_interpretability}
To evaluate the interpretability of the intermediate results, we extract the intermediate values generated by Capacity networks from the trained models and show which intermediate values are expected. Table~\ref{tab:interpretability} illustrates the improved interpretability of Capacity networks in the \mnistredshort{} dataset. It can be seen that the networks learned the underlying set functions reasonably well. In this example, the \rnnc{} generates very good intermediate results while the intermediate results generated by the \gruc{} are less precise - an insight that cannot be gained by investigating the set-level predictions alone.

\subsection{Multiplication-based Problem}
In general, Capacities and Capacity networks can be applied to any (monotone) non-additive problem, which also includes non-additive problems that have less obvious added values such as multiplication-based problems. To illustrate this, we created a multiplication-based dataset according to $f_{\text{\mnistmulshort{}}} (\bag) = \prod_{i=1}^n c_i$, where $c_i$ indicates the class index of instance $i$ and define the empty set to have a utility of 1. Table~\ref{tab:interpretability_prod} shows an illustration of the learned intermediate results to demonstrate that the Capacity networks are able to learn added values as expected.

\begin{table}
	\centering
	\begin{tabular}{r c c c r}
		\multicolumn{5}{c}{ } \\
		Input &
		\scalebox{0.40}{\includegraphics{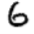}} &
		\scalebox{0.40}{\includegraphics{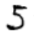}} &
		\scalebox{0.40}{\includegraphics{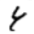}} & \multicolumn{1}{c}{$\sum$} \\
		
		Expected & 6.00    & 24.00    & 90.00    & 120.00   \\
		\midrule
		\multicolumn{1}{c}{\rnnc{}}    & 5.69 & 24.49 & 89.27 & 119.45\\
		\multicolumn{1}{c}{\lstmc{}}   & 5.34 & 25.03 & 90.34 & 120.71\\
		\multicolumn{1}{c}{\gruc{}}    & 2.85 & 26.21 & 88.37 & 117.68\\
	\end{tabular}
	\caption{Intermediate results in a multiplication-based dataset. In this example, the second added value needs to be $24$, since this is the amount that needs to be added to $6$ to obtain $6 \cdot 5=30$.}
	\label{tab:interpretability_prod}
\end{table}

\subsection{Evaluation Beyond Input-Output Testing}
\label{sec:intermediate_result_evaluation}

\begin{table}
	\centering
	\begin{tabular}{l c c c c}
		& \mnistredshort{} & \mnistsynshort{} & \mnistredsynshort{} & \fmnisttrishort{} \\
		\toprule
		\rnnc{} &	\textbf{0.29}	&	0.14	&	\textbf{0.40}	&	1.32\\
		\lstmc{}&	1.50	&	0.09	&	2.00	&	\textbf{0.57}\\
		\gruc{} &	2.15	&	\textbf{0.08}	&	0.47	&	0.62\\
	\end{tabular}
	\caption{Quantitative analysis of the generated intermediate results. We report the mean absolute distance to the expected intermediate results.}
	\label{tab:intermediate_utility_errors_capacity_networks}
\end{table}

Capacity networks also allow in-depth evaluation beyond mere input-output testing. More specifically, for Capacity networks, it is not only possible to evaluate if the model produces the correct final outputs, but also to perform a quantitative evaluation of the intermediate results. The results of this evaluation can be found in Table~\ref{tab:intermediate_utility_errors_capacity_networks}. We find that the obtained error for the set labels does not always reliably indicate which model performs best at predicting intermediate results for individual instances. Hence, the quantitative evaluation provides additional insights, which cannot be obtained for encoder-decoder architectures.

\subsection{Regularizing Latent Intermediate Results}
\label{sec:regularization_experiment}
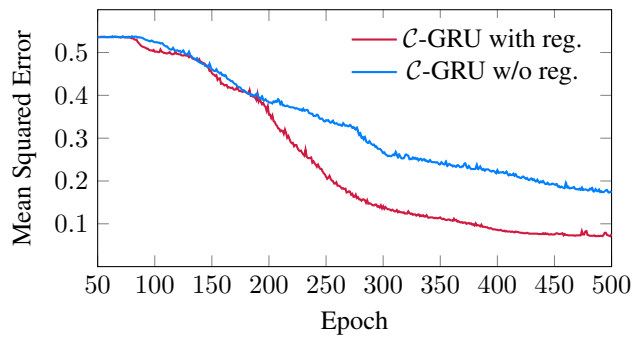
\begin{figure}
	\begin{tikzpicture}
		\begin{axis}[
			width=\linewidth,
			height=5cm,
			legend pos=north east,
			legend style={draw=none},
			xlabel=Epoch,
			ylabel=Mean Squared Error,
			xlabel near ticks,
			ylabel near ticks,
			xmin=50,
			xmax=500,
			ymin=0,
			ymax=0.6,
			xtick={50,100,...,500},
			ytick={0.1,0.2,...,0.5}]
			
			\addplot[no markers, thick, color=nice_red] table[x index=0, y index=1] {results/regularization_learning_curves.txt};
			\addlegendentry{\gruc{} with reg.}
			
			\addplot[no markers, thick, smooth, color=nice_blue] table[x index=0, y index=2] {results/regularization_learning_curves.txt};
			\addlegendentry{\gruc{} w/o reg.}
		\end{axis}
	\end{tikzpicture}
	\vspace{-20pt}
	\caption{Mean squared test error of two \gruc{}s during training. Red and blue lines indicate the learning curves of a \gruc{} with regularization and without regularization, respectively.}
	\label{fig:regularization_experiment}
\end{figure}

In Figure~\ref{fig:regularization_experiment}, we demonstrate how prior domain knowledge can be used in Capacity networks by adding additional regularization on the generated intermediate results. To this end, we use the \fmnistuniqecount{} problem \cite{Murphy2019}. In this task, we can make use of the prior knowledge that no individual contribution can be larger than $1$ by adding a regularization term to the network such that intermediate values larger than 1 are penalized. We find that the added  intermediate results regularization improves the training process and leads to better results in fewer iterations and a better performance after training. Adding this kind of regularization is only possible because Capacity networks produce meaningful intermediate results during training and is not possible for their non-Capacity counterparts and other encoder-decoder architectures. 

\section{Conclusions}
\label{sec:conclusion}
We present Capacity networks, a new family of neural networks to aggregate information from multiple instances. Inspired by non-additive capacities and a sequential decomposition thereof, Capacity networks produce a latent intermediate result for each instance, which models the added value of the instance with respect to all already observed instances. This aggregation strategy differs fundamentally from previously used encoder-decoder architectures. Our experiments show that Capacity networks systematically outperform their non-capacity counterparts and other prior architectures, which do not produce intermediate results, in a wide range of setups. Furthermore, we demonstrate the improved interpretability of Capacity networks that allows a detailed inspection of the network internals, an evaluation beyond mere input-output testing, and incorporation of prior knowledge via intermediate result regularization.

\section{Future Work}
\label{sec:future_work}
It is noteworthy that inductive bias presented in this work is not limited to MIL. Aggregating information from multiple sources is also a key problem in areas such as multi-modal learning and geometric deep learning. Hence, exploring more application areas beyond MIL for Capacity networks is promising. Furthermore, extending Capacity networks to classification problems is a promising future research direction, especially if the monotonicity constrained is not enforced. Moreover, prior works have shown that learning to order instances can further improve network accuracy \cite{Vinyals2016}. In future research, it would be interesting to see if and to which extent this also applies to Capacity networks.

\bibliography{references}

\begin{thebibliography}{49}
\providecommand{\natexlab}[1]{#1}

\bibitem[{Anderson et~al.(2018)Anderson, Grant, Islam, Murray, and
  Marcum}]{Anderson2018}
Anderson, D.~T.; Grant, J.~S.; Islam, M.~A.; Murray, B.; and Marcum, R. 2018.
\newblock {Fuzzy Choquet Integration of Deep Convolutional Neural Networks for
  Remote Sensing}.
\newblock In \emph{Computational Intelligence for Pattern Recognition}, 1--28.
  Springer International Publishing.

\bibitem[{Bahdanau, Cho, and Bengio(2015)}]{Bahdanau2015}
Bahdanau, D.; Cho, K.~H.; and Bengio, Y. 2015.
\newblock {Neural Machine Translation by Jointly Learning to Align and
  Translate}.
\newblock In \emph{Proceeding of the 3rd International Conference on Learning
  Representations}, 1--15.

\bibitem[{Beck et~al.(2020)Beck, Ciosek, Devlin, Tschiatschek, Zhang, and
  Hofmann}]{Beck2020}
Beck, J.; Ciosek, K.; Devlin, S.; Tschiatschek, S.; Zhang, C.; and Hofmann, K.
  2020.
\newblock {AMRL: Aggregated Memory For Reinforcement Learning}.
\newblock In \emph{Proceedings of the 8th International Conference on Learning
  Representations}, 1--14.

\bibitem[{Beliakov(2008)}]{Beliakov2008}
Beliakov, G. 2008.
\newblock {Fitting fuzzy measures by linear programming. Programming library
  fmtools}.
\newblock \emph{IEEE International Conference on Fuzzy Systems}, 862--867.

\bibitem[{Carbonneau et~al.(2018)Carbonneau, Cheplygina, Granger, and
  Gagnon}]{Carbonneau2018}
Carbonneau, M.~A.; Cheplygina, V.; Granger, E.; and Gagnon, G. 2018.
\newblock {Multiple instance learning: A survey of problem characteristics and
  applications}.
\newblock \emph{Pattern Recognition}, 77: 329--353.

\bibitem[{Cho et~al.(2014)Cho, van Merri{\"{e}}nboer, Gulcehre, Bahdanau,
  Bougares, Schwenk, and Bengio}]{Cho2014}
Cho, K.; van Merri{\"{e}}nboer, B.; Gulcehre, C.; Bahdanau, D.; Bougares, F.;
  Schwenk, H.; and Bengio, Y. 2014.
\newblock {Learning Phrase Representations using RNN Encoder–Decoder for
  Statistical Machine Translation}.
\newblock In \emph{Proceedings of the 2014 Conference on Empirical Methods in
  Natural Language Processing}, 1724--1734.

\bibitem[{Choquet(1954)}]{Choquet1954}
Choquet, G. 1954.
\newblock {Theory of capacities}.
\newblock \emph{Annales de l'Institut Fourier}, 5: 131--295.

\bibitem[{Das, Giles, and Sun(1992)}]{Das1992}
Das, S.; Giles, C.~L.; and Sun, G.-Z. 1992.
\newblock {Learning Context-free Grammars: Capabilities and Limitations of a
  Recurrent Neural Network with an External Stack Memory}.
\newblock In \emph{Proceedings of the 14th Annual Conference of the Cognitive
  Science Society}, 791--795.

\bibitem[{Dias et~al.(2018)Dias, Bueno, Borges, Botelho, Dimuro, , Lucca,
  Fernand{\'{e}}z, Bustince, and {Drews Junior}}]{Dias2018}
Dias, C.~A.; Bueno, J. C.~S.; Borges, E.~N.; Botelho, S. S.~C.; Dimuro, G.~P.;
  ; Lucca, G.; Fernand{\'{e}}z, J.; Bustince, H.; and {Drews Junior}, P. L.~J.
  2018.
\newblock {Using the Choquet Integral in the Pooling Layer in Deep Learning
  Networks}.
\newblock In \emph{Fuzzy Information Processing}, 144--154. Springer
  International Publishing.

\bibitem[{Dietterich, Lathrop, and Lozano-P{\'{e}}rez(1997)}]{Dietterich1997}
Dietterich, T.~G.; Lathrop, R.~H.; and Lozano-P{\'{e}}rez, T. 1997.
\newblock {Solving the multiple instance problem with axis-parallel
  rectangles}.
\newblock \emph{Artificial Intelligence}, 89(1-2): 31--71.

\bibitem[{Graves, Wayne, and Danihelka(2014)}]{Graves2014}
Graves, A.; Wayne, G.; and Danihelka, I. 2014.
\newblock {Neural Turing Machines}.
\newblock In \emph{arXiv preprint}, 1--26.

\bibitem[{Herbrich, Minka, and Graepel(2006)}]{Herbrich2006TrueSkill}
Herbrich, R.; Minka, T.; and Graepel, T. 2006.
\newblock {TrueSkill: A Bayesian Skill Rating System}.
\newblock In \emph{Advances in Neural Information Processing Systems 20},
  569--576.

\bibitem[{Hochreiter and Schmidhuber(1997)}]{Hochreiter1997}
Hochreiter, S.; and Schmidhuber, J. 1997.
\newblock {Long Short-Term Memory}.
\newblock \emph{Neural Computation}, 9(8): 1735--1780.

\bibitem[{Ilse, Tomczak, and Welling(2018)}]{Ilse2018}
Ilse, M.; Tomczak, J.~M.; and Welling, M. 2018.
\newblock {Attention-based deep multiple instance learning}.
\newblock In \emph{Proceedings of the 35th International Conference on Machine
  Learning}, 3376--3391.
\newblock ISBN 9781510867963.

\bibitem[{Kalra et~al.(2020)Kalra, Adnan, Taylor, and Tizhoosh}]{Kalra2020}
Kalra, S.; Adnan, M.; Taylor, G.; and Tizhoosh, H. 2020.
\newblock {Learning Permutation Invariant Representations using Memory
  Networks}.
\newblock In \emph{Proceedings of the 16th European Conference on Computer
  Vision}.

\bibitem[{Karczmarek, Kiersztyn, and Pedrycz(2018)}]{Karczmarek2018}
Karczmarek, P.; Kiersztyn, A.; and Pedrycz, W. 2018.
\newblock {Generalized Choquet Integral for Face Recognition}.
\newblock \emph{International Journal of Fuzzy Systems}, 20(3): 1047--1055.

\bibitem[{Karczmarek, Kiersztyn, and Pedrycz(2019)}]{Karczmarek2019}
Karczmarek, P.; Kiersztyn, A.; and Pedrycz, W. 2019.
\newblock {Generalizations of Aggregation Functions for Face Recognition}.
\newblock In \emph{Artificial Intelligence and Soft Computing}, 182--192.

\bibitem[{Kingma and Ba(2015)}]{Kingma2015}
Kingma, D.~P.; and Ba, J.~L. 2015.
\newblock {Adam: A method for stochastic optimization}.
\newblock In \emph{Proceedings of the 3rd International Conference on Learning
  Representations}, 1--15.

\bibitem[{LeCun et~al.(1998)LeCun, Bottou, Bengio, and Haffner}]{LeCun1998}
LeCun, Y.; Bottou, L.; Bengio, Y.; and Haffner, P. 1998.
\newblock {Gradient-Based Learning Applied to Document Recognition}.
\newblock \emph{Proceedings of the IEEE}, 86(11): 2278--2324.

\bibitem[{Lee et~al.(2019)Lee, Lee, Kim, Kosiorek, Choi, and Teh}]{Lee2019}
Lee, J.; Lee, Y.; Kim, J.; Kosiorek, A.~R.; Choi, S.; and Teh, Y.~W. 2019.
\newblock {Set Transformer: A Framework for Attention-based
  Permutation-Invariant Neural Networks}.
\newblock In \emph{Proceedings of the 36th International Conference on Machine
  Learning}, 3744--3753.

\bibitem[{Luong, Pham, and Manning(2015)}]{Luong2015}
Luong, M.-T.; Pham, H.; and Manning, C.~D. 2015.
\newblock {Effective Approaches to Attention-based Neural Machine Translation}.
\newblock In \emph{Proceedings of the 2015 Conference on Empirical Methods in
  Natural Language Processing}, 1412--1421.

\bibitem[{Maron and Lozano-Perez(1998)}]{Maron1998}
Maron, O.; and Lozano-Perez, T. 1998.
\newblock {A framework for multiple-instance learning}.
\newblock \emph{Advances in Neural Information Processing Systems}, 570--576.

\bibitem[{McAuley, Leskovec, and Jurafsky(2012)}]{McAuley2012}
McAuley, J.; Leskovec, J.; and Jurafsky, D. 2012.
\newblock {Learning attitudes and attributes from multi-aspect reviews}.
\newblock \emph{Proceedings of the 12th IEEE International Conference on Data
  Mining}, 1020--1025.

\bibitem[{Mena et~al.(2018)Mena, Belanger, Linderman, and Snoek}]{Mena2018}
Mena, G.; Belanger, D.; Linderman, S.; and Snoek, J. 2018.
\newblock {Learning Latent Permutations with Gumbel-Sinkhorn Networks}.

\bibitem[{Meng et~al.(2019)Meng, Yang, Ribeiro, and Neville}]{Meng2019}
Meng, C.; Yang, J.; Ribeiro, B.; and Neville, J. 2019.
\newblock {HATS: A hierarchical sequence-attention framework for inductive
  set-of-sets embeddings}.
\newblock \emph{Proceedings of the ACM SIGKDD International Conference on
  Knowledge Discovery and Data Mining}, 783--792.

\bibitem[{Murphy et~al.(2019)Murphy, Srinivasan, Ribeiro, and Rao}]{Murphy2019}
Murphy, R.~L.; Srinivasan, B.; Ribeiro, B.; and Rao, V. 2019.
\newblock {Janossy pooling: Learning deep permutation-invariant functions for
  variable-size inputs}.
\newblock In \emph{Proceedings of the 7th International Conference on Learning
  Representations}, 1--21.

\bibitem[{Niepert, Ahmed, and Kutzkov(2016)}]{Niepert2016}
Niepert, M.; Ahmed, M.; and Kutzkov, K. 2016.
\newblock {Learning Convolutional Neural Networks for Graphs}.
\newblock In \emph{Proceedings of the 33rd International Conference on Machine
  Learning}, 2014--2023.

\bibitem[{Pappas and Popescu-Belis(2017)}]{Pappas2017}
Pappas, N.; and Popescu-Belis, A. 2017.
\newblock {Explicit document modeling through weighted multiple-instance
  learning}.
\newblock \emph{Journal of Artificial Intelligence Research}, 58: 591--626.

\bibitem[{Qi et~al.(2017)Qi, Su, Mo, and Guibas}]{Qi2017}
Qi, C.~R.; Su, H.; Mo, K.; and Guibas, L.~J. 2017.
\newblock {PointNet: Deep learning on point sets for 3D classification and
  segmentation}.
\newblock \emph{Proceedings of the IEEE Conference on Computer Vision and
  Pattern Recognition}, 77--85.

\bibitem[{Ray and Page(2001)}]{Ray2001}
Ray, S.; and Page, D. 2001.
\newblock {Multiple Instance Regression}.
\newblock In \emph{Proceedings of the International Conference on Machine
  Learning}, 425 -- 432.

\bibitem[{Schmidhuber(1992)}]{Schmidhuber1992}
Schmidhuber, J. 1992.
\newblock {Learning to Control Fast-weight Memories: an Alternative to Dynamic
  Recurrent Networks}.
\newblock \emph{Neural Computation}, 4(1): 131--139.

\bibitem[{Schmidhuber(1993)}]{Schmidhuber1993}
Schmidhuber, J. 1993.
\newblock {A ‘Self-Referential' Weight Matrix}.
\newblock In \emph{Proceedings of the International Conference on Artificial
  Neural Networks}, 446--450.

\bibitem[{Schmidt, Schneider, and Hennig(2020)}]{Schmidt2020}
Schmidt, R.~M.; Schneider, F.; and Hennig, P. 2020.
\newblock {Descending through a Crowded Valley - Benchmarking Deep Learning
  Optimizers}.
\newblock In \emph{arXiv}, 1--30.

\bibitem[{Scott et~al.(2017)Scott, Marcum, Davis, and Nivin}]{Scott2017}
Scott, G.~J.; Marcum, R.~A.; Davis, C.~H.; and Nivin, T.~W. 2017.
\newblock {Fusion of Deep Convolutional Neural Networks for Land Cover
  Classification of High-Resolution Imagery}.
\newblock \emph{IEEE Geoscience and Remote Sensing Letters}, 14(9): 1638--1642.

\bibitem[{Segol and Lipman(2020)}]{Segol2020}
Segol, N.; and Lipman, Y. 2020.
\newblock {On Universal Equivariant Set Networks}.
\newblock In \emph{Proceedings of the 8th International Conference on Learning
  Representations}.

\bibitem[{Shi, Oliva, and Niethammer(2020)}]{Shi2020}
Shi, Y.; Oliva, J.; and Niethammer, M. 2020.
\newblock {Deep Message Passing on Sets}.
\newblock In \emph{Proceedings of the 34th AAAI Conference on Artificial
  Intelligence}, 5750--5757.

\bibitem[{Soelch et~al.(2019)Soelch, Akhundov, van~der Smagt, and
  Bayer}]{Soelch2019}
Soelch, M.; Akhundov, A.; van~der Smagt, P.; and Bayer, J. 2019.
\newblock {On Deep Set Learning and the Choice of Aggregations}.
\newblock In \emph{Proceeding of the 28th International Conference on
  Artificial Neural Networks}, 444--457. Springer International Publishing.

\bibitem[{Sudharshana et~al.(2019)Sudharshana, Petitjean, Spanhol, Oliveira,
  Honeine, Sudharshana, Petitjean, Spanhol, Oliveira, and
  Heutte}]{Sudharshana2019}
Sudharshana, P.~J.; Petitjean, C.; Spanhol, F.; Oliveira, L.; Honeine, P.;
  Sudharshana, P.~J.; Petitjean, C.; Spanhol, F.; Oliveira, L.; and Heutte, L.
  2019.
\newblock {Multiple Instance Learning for Histopathological Breast Cancer
  Images}.
\newblock \emph{Expert Systems with Applications}, 117: 103--111.

\bibitem[{Sugeno(1974)}]{Sugeno1974}
Sugeno, M. 1974.
\newblock \emph{{Theory of fuzzy integrals and its applications}}.
\newblock Ph.d. thesis, Tokyo Institute of Technology.

\bibitem[{Tehrani et~al.(2012)Tehrani, Cheng, Dembczy{\'{n}}ski, and
  H{\"{u}}llermeier}]{Tehrani2012a}
Tehrani, A.~F.; Cheng, W.; Dembczy{\'{n}}ski, K.; and H{\"{u}}llermeier, E.
  2012.
\newblock {Learning monotone nonlinear models using the Choquet integral}.
\newblock \emph{Machine Learning}, 89(1-2): 183--211.

\bibitem[{Tehrani, Cheng, and Hullermeier(2012)}]{Tehrani2012}
Tehrani, A.~F.; Cheng, W.; and Hullermeier, E. 2012.
\newblock {Preference learning using the choquet integral: The case of
  multipartite ranking}.
\newblock \emph{IEEE Transactions on Fuzzy Systems}, 20(6): 1102--1113.

\bibitem[{Vaswani et~al.(2017)Vaswani, Shazeer, Parmar, Uszkoreit, Jones,
  Gomez, Kaiser, and Polosukhin}]{Vaswani2017}
Vaswani, A.; Shazeer, N.; Parmar, N.; Uszkoreit, J.; Jones, L.; Gomez, A.~N.;
  Kaiser, L.; and Polosukhin, I. 2017.
\newblock {Attention Is All You Need}.
\newblock In \emph{Advances in Neural Information Processing Systems 30},
  6000--6010.

\bibitem[{Vinyals, Bengio, and Kudlur(2016)}]{Vinyals2016}
Vinyals, O.; Bengio, S.; and Kudlur, M. 2016.
\newblock {Order matters: Sequence to sequence for sets}.
\newblock In \emph{Proceeding of the 4th International Conference on Learning
  Representations}, 1--11.

\bibitem[{Weston, Chopra, and Bordes(2015)}]{Weston2015}
Weston, J.; Chopra, S.; and Bordes, A. 2015.
\newblock {Memory Networks}.
\newblock In \emph{Proceeding of the 3rd International Conference on Learning
  Representations}, 1--15.

\bibitem[{Xiao, Rasul, and Vollgraf(2017)}]{Xiao2017}
Xiao, H.; Rasul, K.; and Vollgraf, R. 2017.
\newblock {Fashion-MNIST: A novel image dataset for benchmarking machine
  learning algorithms}.
\newblock In \emph{arXiv preprint}, 1--6.

\bibitem[{Yang et~al.(2020)Yang, Wang, Markham, and Trigoni}]{Yang2020a}
Yang, B.; Wang, S.; Markham, A.; and Trigoni, N. 2020.
\newblock {Robust Attentional Aggregation of Deep Feature Sets for Multi-view
  3D Reconstruction}.
\newblock \emph{International Journal of Computer Vision}, 128(1): 53--73.

\bibitem[{Zaheer et~al.(2017)Zaheer, Kottur, Ravanbakhsh, Poczos,
  Salakhutdinov, and Smola}]{Zaheer2017}
Zaheer, M.; Kottur, S.; Ravanbakhsh, S.; Poczos, B.; Salakhutdinov, R.; and
  Smola, A. 2017.
\newblock {Deep Sets}.
\newblock In \emph{Proceeding of the 31st Conference on Neural Information
  Processing Systems}, 1--11.

\bibitem[{Zhang, Hare, and Pr{\"{u}}gel-Bennett(2020)}]{Zhang2020}
Zhang, Y.; Hare, J.; and Pr{\"{u}}gel-Bennett, A. 2020.
\newblock {FSPool: Learning Set Representations with Featurewise Sort Pooling}.
\newblock In \emph{Proceedings of the 8th International Conference on Learning
  Representations}.

\bibitem[{Zhang, Pr{\"{u}}gel-Bennett, and Hare(2019)}]{Zhang2019}
Zhang, Y.; Pr{\"{u}}gel-Bennett, A.; and Hare, J. 2019.
\newblock {Learning Representations of Sets through Optimized Permutations}.
\newblock In \emph{Proceeding of the 7th International Conference on Learning
  Representations}, 1--15.

\end{thebibliography}

\clearpage

\vbox{\centering {\LARGE\bf Supplementary Material}}

\appendix

In the supplementary material, we provide additional information on implementation details, model sizes, and the hyperparameter optimization. Furthermore, we provide additional results for experiments in the main paper and generate and evaluate potential intermediate results of non-Capacity networks.

\section{Implementation Details}
\label{sec:app_implementation_details}
In the following, we provide additional information on the different architectures used in the experiments. As already described in the main paper, the architectures can be described in terms of encoder and decoder modules. Different conventions are used to label different parts of the networks. In particular, the pooling operation used in parallel networks can be viewed as part of the encoder or as additional module. As a consequence, these networks have also been described as encoder-pooling-decoder networks \cite{Lee2019}. In this work, we consider the pooling layer as part of the encoder, since the pooling generates a joint representation of all input instances, a task that belongs to the domain of encoders. Similar to \citet{Zaheer2017}, we performed initial experiments with 2 and 3 layer fully connected networks for the encoder and decoder modules (see Appendix~\ref{sec:app_hyperparameter_optimization} for more information). However, using more complex networks is possible and perhaps beneficial for more complex input data. We denote the fully connected layers in the encoder and decoder by $\epsilon$ ('epsilon' for 'encoder') and $\delta$ ('delta' for 'decoder'), respectively. Furthermore, we note that the networks only use one instantiation of $\epsilon$ and $\delta$. Hence, different applications of $\epsilon$ and $\delta$ in the architectures share parameters with all other applications.

All models use ReLu activation in the encoders and decoders. We have also performed initial experiments with Sigmoid and Tanh activation functions. However, we observed that ReLu often performs better in terms of mean squared error and results in a more stable training process. ReLu activations have also been used by prior works \cite{Zaheer2017}.

In non-Capacity networks, we denote the set representation by $\latsetrep$, whereas in Capacity networks, we denote the representation used to generate the intermediate outputs by $\latinstancerep_i$. All models use the same number of dimensions for internal representation (i.e. $\size{\latsetrep} = \size{\latinstancerep_i}$). Having the same number of dimensions further improves the comparability of the models, since the performance differences between the models cannot be explained by different hidden dimension sizes. 

In the following, we describe further architecture-specific details.

\paragraph{\dset{}}
\begin{equation}
	\begin{split}
		& Z = \encoder(X) = \sum_{x \in X} \epsilon(x) \\
		& Y = \decoder(\latsetrep) = \delta(\latsetrep) \\
	\end{split}
\end{equation}

\paragraph{\att{}}
\begin{equation}
	\begin{split}
		& Z = \encoder(X) = \sum_{x \in X} a_i \cdot \epsilon(x_i) \\
		& a_i = \frac{\exp{h(\instance_i)}}{\sum_{j=1}^{\size{X}} \exp{h(\instance_j)}} \\
		& h(x) = B \cdot \tanh(A \cdot x + a) + b, \\
		& A \in \reals{}^{n \times d}, B \in \reals{}^{1 \times n}, a \in \reals{}^{n}, b \in \reals{} \\
		& Y = \decoder(\latsetrep) = \delta(\latsetrep) \\
	\end{split}
\end{equation}

\paragraph{\rnn{}}
\begin{equation}
	\begin{split}
		& Z = \encoder(X) = h_n \\
		& h_i = \tanh(A \cdot [h_{i-1}, x_i] + a), \\
		& A \in \reals^{2n \times d}, a \in \reals^{d} \\
		& Y = \decoder(\latsetrep) = \delta(\latsetrep) \\
	\end{split}
\end{equation}
where $[\cdot,\cdot]$ denotes concatenation of two vectors, $d$ equals the number of hidden dimensions, and $h_0 = \vec{0}$ is used as initial hidden state.

\paragraph{\lstm{}}
\begin{equation}
	\begin{split}
		& Z = \encoder(X) = \text{LSTM}_i \\
		& Y = \decoder(\latsetrep) = \delta(\latsetrep) \\
	\end{split}
\end{equation}
where $\text{LSTM}_i$ is an LSTM cell according to \citet{Hochreiter1997} and $c_0 = \vec{0}$ is used as initial cell state.

\paragraph{\gru{}}
\begin{equation}
	\begin{split}
		& Z = \encoder(X) = \text{GRU}_i \\
		& Y = \decoder(\latsetrep) = \delta(\latsetrep) \\
	\end{split}
\end{equation}
where $\text{GRU}_i$ is a GRU cell according to \citet{Cho2014} and $c_0 = \vec{0}$ is used as initial cell state.

\paragraph{\rnnc{}}
\begin{equation}
	\begin{split}
		& \latinstancerep_i = \encoder(x_i) = h_i \\
		& h_i = \tanh(A \cdot [h_{i-1}, x_i] + a), \\
		& A \in \reals^{2n \times d}, a \in \reals^{d}, h_0 = \vec{0} \\
		& \latobjutility_i = \decoder(\latinstancerep_i) = \delta(\latinstancerep_i) \\
		& Y = \sum_{i=0}^{\size{X}} \latobjutility_i
	\end{split}
\end{equation}

\paragraph{\lstmc{}}
\begin{equation}
	\begin{split}
		& \latinstancerep_i = \encoder(x_i) = \text{LSTM}_i \\
		& \latobjutility_i = \decoder(\latinstancerep_i) = \delta(\latinstancerep_i) \\
		& Y = \sum_{i=0}^{\size{X}} \latobjutility_i
	\end{split}
\end{equation}

\paragraph{\gruc{}}
\begin{equation}
	\begin{split}
		& \latinstancerep_i = \encoder(x_i) = \text{GRU}_i \\
		& \latobjutility_i = \decoder(\latinstancerep_i) = \delta(\latinstancerep_i) \\
		& Y = \sum_{i=0}^{\size{X}} \latobjutility_i
	\end{split}
\end{equation}

Furthermore, we refer to \citet{Lee2019} and to \url{http://github.com/juho-lee/set_transformer} for a detailed description and implementation of the Set Transformer architecture.

\clearpage
\section{Model Sizes}
\label{sec:app_model_sizes}
Table~\ref{tab:number_of_parameters} shows the number of trainable parameters for each model. Most important, all sequential models (i.e. \rnn{}, \lstm{}, and \gru{}) and their Capacity Network counterparts (i.e. \rnnc{}, \lstmc{}, and \gruc{}) have the same number of parameters. Hence, the performance differences observed in the experiments cannot be explained by different model sizes.

\begin{table}[h!]
	\centering
	\begin{tabular}{l r}
		Architecture & Parameters \\
		\toprule
		\dset{}	&	13,601\\
		\att{}	&	17,794\\
		\smallst{}&	15,809\\
		\st{}		&	34,753\\
		\lightmidrule									
		\rnn{}	&	15,681\\
		\lstm{}	&	22,049\\
		\gru{}	&	19,937\\
		\lightmidrule									
		\rnnc{} (ours)	&	15,681\\
		\lstmc{} (ours)	&	22,049\\
		\gruc{} (ours)	&	19,937\\
		\bottomrule
	\end{tabular}
	\caption{Number of trainable parameters for instance embedding size of 64 and latent representation dimension of 32.}
	\label{tab:number_of_parameters}
\end{table}

\section{Hyperparameter Optimization}
\label{sec:app_hyperparameter_optimization}
We performed a hyperparameter search for the following parameters and ranges: learning rate $\in \set{0.01, 0.001, 0.0001}$, number of hidden dimensions $\in \set{8,16,32}$, number of fully connected layers for encoders and decoders $\in \set{2,3}$. In total, we performed 486 experiments for 10 architectures and 3 datasets. Table~\ref{tab:hyperparameter_optimization} shows the results of the hyperparameter optimization. It can be seen that all architectures perform better with 3 instead of 2 fully connected layers for the encoder and decoder sub-networks. Moreover, 32 dimensional latent representations work best for most networks. The only exception are $\gruc{}$s, for which 16 dimensions work equally well. Furthermore, a learning rate of $0.001$ leads to best results for all networks. Learning rates of $0.01$ and $0.0001$ lead to good results in only 2 out of 30 setups. To sum, we find a common trend for all networks and use a learning rate of $0.001$, $32$ hidden dimensions, and $3$ fully connected layers in all experiments. Furthermore, we performed experiments with different batch sizes in the range of $100$, $500$, $1000$, and $2000$ and found that a batch size of $1000$ works well in general.

\section{Dataset Statistics}
\label{sec:app_dataset_statistics}
In this section, we report statistics of the generated datasets. Table~\ref{tab:dataset_statistics} shows mean, median, variance, and standard deviation of several datasets. Figure~\ref{fig:set_utility_distributions} illustrates the distribution of set utilities for the datasets \mnistredshort{}, \mnistsynshort{}, and \mnistredsynshort.

\begin{table}
	\centering
	\begin{tabular}{l c r r r r}
		Dataset & Set Size & Mean & Med & Variance & StDev \\
		\toprule
		\mnistredshort{} & 10 & 29.34  & 30  & 39.88  & 6.32 \\
		\mnistsynshort{} & 10 & 65.19  & 62  & 407.16  & 20.18 \\
		\mnistredsynshort{} & 10 & 49.80  & 50  & 169.87  & 13.03 \\
		\fmnisttrishort{} & 10 & 14.50  & 14  & 4.02  & 2.00 \\
		\lightmidrule
		\mnistsynshort{} & 20 & 175.42 & 170.0 & 1869.67 & 43.24 \\
		\mnistsynshort{} & 30 & 330.77 & 323.0 & 4885.71 & 69.9 \\
		\mnistsynshort{} & 40 & 530.93 & 521.0 & 9805.65 & 99.02 \\
		\lightmidrule
		\mnistredshort{} & var. & 28.58  & 29  & 60.73  & 7.79 \\
		\mnistsynshort{} & var. & 66.84  & 62  & 1068.33  & 32.69 \\
		\mnistredsynshort{} & var. & 50.56  & 50  & 292.40  & 17.10 \\
		\fmnisttrishort{} & var. & 32.11  & 29  & 108.26  & 10.40 \\
		\bottomrule
	\end{tabular}
	\caption{Mean, median (Med), variance, and standard deviation (StDev) of generated datasets with fixed and varying set sizes.}
	\label{tab:dataset_statistics}
\end{table}

\section{Additional Experimental Results for Table~\ref{tab:full_training_results}}
Table~\ref{tab:full_training_results_median} and Table~\ref{tab:full_training_results_stdev} provide further details for the experiments in Table~\ref{tab:full_training_results}. We report median and standard deviation to allow a better interpretation of the reported results.
\begin{table}[h!]
	\centering
	\begin{tabular}{l r r r r}
		& \multicolumn{1}{c}{\mnistredshort{}} &  \multicolumn{1}{c}{\mnistsynshort{}} & \multicolumn{1}{c}{\mnistredsynshort{}} & \multicolumn{1}{c}{\fmnisttrishort{}} \\
		\toprule							
		\rnn	&	0.91	&	1.65	&	7.04	&	4.09	\\
		\lstm	&	0.41	&	0.87	&	2.80	&	\textbf{1.01}	\\
		\gru	&	0.48	&	0.89	&	2.56	&	3.98	\\
		\lightmidrule									
		\rnnc{} (ours)	&	0.44	&	0.61	&	2.61	&	4.08	\\
		\lstmc{} (ours)	&	0.27	&	0.56	&	\textbf{1.62}	&	\textbf{1.01}	\\
		\gruc{} (ours)	&	0.27	&	\textbf{0.48}	&	1.93	&	1.26	\\
		\lightmidrule	
		\dset	&	\textbf{0.24}	&	2.87	&	2.43	&	1.02	\\
		\att	&	0.53	&	0.94	&	3.01	&	4.08	\\
		\smallst&	1.93	&	8.39	&	15.15	&	1.21	\\
		\st		&	1.66	&	2.80	&	11.31	&	1.19	\\
		
		\bottomrule
	\end{tabular}
	\caption{Median of the results in Table~\ref{tab:full_training_results}.}
	\label{tab:full_training_results_median}
\end{table}

\begin{table}[h!]
	\centering
	\begin{tabular}{l r r r r}
		& \multicolumn{1}{c}{\mnistredshort{}} &  \multicolumn{1}{c}{\mnistsynshort{}} & \multicolumn{1}{c}{\mnistredsynshort{}} & \multicolumn{1}{c}{\fmnisttrishort{}} \\
		\toprule							
		\rnn	&	0.10	&	0.21	&	0.64	&	0.00	\\
		\lstm	&	0.03	&	0.21	&	0.18	&	0.01	\\
		\gru	&	0.04	&	0.24	&	0.29	&	1.44	\\
		\lightmidrule									
		\rnnc{} (ours)	&	0.03	&	0.07	&	0.21	&	1.30	\\
		\lstmc{} (ours)	&	0.01	&	0.13	&	0.03	&	0.02	\\
		\gruc{} (ours)	&	0.06	&	0.08	&	0.49	&	1.39	\\
		\lightmidrule
		\dset	&	0.06	&	0.15	&	0.52	&	1.47	\\
		\att	&	0.20	&	0.21	&	0.96	&	1.36	\\
		\smallst	&	0.09	&	1.44	&	1.25	&	0.02	\\
		\st	&	0.11	&	0.61	&	2.47	&	0.02	\\
		\bottomrule
	\end{tabular}
	\caption{Standard deviation (StDev) for the experiments in Table~\ref{tab:full_training_results}.}
	\label{tab:full_training_results_stdev}
\end{table}

\begin{table*}
	\footnotesize
	\centering
	\begin{tabular}{l r r r | r r r | r r r | r r r}
		\multicolumn{4}{c}{ } & \multicolumn{3}{c}{\mnistredshort{}} & \multicolumn{3}{c}{\mnistsynshort{}} & \multicolumn{3}{c}{\mnistredsynshort{}} \\
		\cmidrule(lr){5-7} \cmidrule(lr){8-10}  \cmidrule(lr){11-13}
		Network & E/D & ND & LR= & 0.01 &  0.001 &  0.0001 & 0.01 &  0.001 &  0.0001 & 0.01 &  0.001 &  0.0001 \\
		\midrule
		
		\dset{}	&	2fc	&	8	&		&	10.53	&	4.90	&	10.45	&	40.09	&	33.65	&	101.69	&	111.83	&	43.50	&	112.10	\\
		&		&	16	&		&	6.43	&	2.73	&	3.35	&	19.06	&	22.66	&	52.38	&	52.13	&	16.07	&	29.37	\\
		&		&	32	&		&	2.35	&	2.19	&	3.10	&	19.80	&	19.15	&	27.72	&	15.31	&	14.14	&	18.34	\\
		&	\textbf{3fc}	&	8	&		&	13.12	&	3.11	&	3.84	&	7.94	&	11.23	&	26.10	&	157.40	&	-	&	88.05	\\
		&		&	16	&		&	0.39	&	0.30	&	3.80	&	\textbf{2.97}	&	5.19	&	15.30	&	5.56	&	-	&	29.25	\\
		&		&	\textbf{32}	&		&	0.53	&	\textbf{0.27}	&	0.94	&	3.20	&	3.60	&	11.86	&	8.68	&	\textbf{2.79}	&	7.16	\\
		\lightmidrule																									
		\att{}	&	2fc	&	8	&		&	-	&	7.20	&	9.54	&	-	&	28.08	&	52.37	&	-	&	40.34	&	78.56	\\
		&		&	16	&		&	-	&	0.88	&	2.89	&	-	&	5.43	&	12.72	&	-	&	6.38	&	37.33	\\
		&		&	32	&		&	-	&	0.52	&	1.25	&	-	&	3.76	&	7.95	&	-	&	3.60	&	11.97	\\
		&	\textbf{3fc}	&	8	&		&	-	&	-	&	-	&	-	&	-	&	-	&	-	&	-	&	-	\\
		&		&	16	&		&	-	&	\textbf{0.21}	&	1.77	&	-	&	2.16	&	5.93	&	-	&	3.80	&	44.75	\\
		&		&	\textbf{32}	&		&	-	&	0.22	&	0.46	&	-	&	\textbf{0.98}	&	4.12	&	-	&	\textbf{1.77}	&	28.30	\\
		\lightmidrule																									
		\rnnc{}	&	2fc	&	8	&		&	-	&	5.63	&	10.49	&	-	&	36.61	&	54.72	&	-	&	83.70	&	90.56	\\
		&		&	16	&		&	-	&	1.38	&	4.16	&	-	&	5.42	&	26.88	&	-	&	18.34	&	63.88	\\
		&		&	32	&		&	-	&	0.80	&	1.63	&	-	&	2.60	&	10.22	&	-	&	6.58	&	15.37	\\
		&	\textbf{3fc}	&	8	&		&	-	&	4.21	&	8.59	&	-	&	25.26	&	36.73	&	-	&	74.67	&	91.49	\\
		&		&	16	&		&	-	&	0.48	&	2.43	&	-	&	1.93	&	14.38	&	-	&	13.47	&	53.18	\\
		&		&	\textbf{32}	&		&	-	&	\textbf{0.47}	&	0.79	&	-	&	\textbf{0.57}	&	6.48	&	-	&	\textbf{2.86}	&	11.99	\\
		\lightmidrule																									
		\lstmc{}	&	2fc	&	8	&		&	-	&	1.39	&	2.97	&	14.09	&	12.09	&	27.01	&	-	&	31.68	&	67.64	\\
		&		&	16	&		&	-	&	0.48	&	1.41	&	-	&	5.10	&	14.16	&	-	&	3.02	&	7.97	\\
		&		&	32	&		&	-	&	0.44	&	0.92	&	-	&	2.38	&	8.21	&	-	&	2.65	&	5.52	\\
		&	\textbf{3fc}	&	8	&		&	-	&	2.22	&	3.95	&	-	&	5.82	&	22.62	&	-	&	26.34	&	53.95	\\
		&		&	16	&		&	-	&	\textbf{0.20}	&	1.38	&	39.67	&	1.12	&	13.08	&	-	&	1.73	&	8.24	\\
		&		&	\textbf{32}	&		&	-	&	0.23	&	0.50	&	-	&	\textbf{0.52}	&	5.02	&	-	&	\textbf{1.04}	&	3.80	\\
		\lightmidrule																									
		\gruc{}	&	2fc	&	8	&		&	-	&	1.53	&	3.86	&	-	&	15.67	&	28.21	&	-	&	29.71	&	66.70	\\
		&		&	16	&		&	-	&	0.58	&	1.25	&	-	&	3.59	&	11.17	&	-	&	3.99	&	8.05	\\
		&		&	32	&		&	-	&	0.47	&	0.63	&	-	&	3.40	&	8.37	&	-	&	3.18	&	4.87	\\
		&	\textbf{3fc}	&	8	&		&	-	&	1.47	&	2.04	&	-	&	9.84	&	20.02	&	-	&	20.28	&	37.06	\\
		&		&	\textbf{16}	&		&	-	&	\textbf{0.29}	&	0.71	&	-	&	0.84	&	7.32	&	-	&	\textbf{1.39}	&	4.85	\\
		&		&	\textbf{32}	&		&	-	&	\textbf{0.29}	&	0.50	&	-	&	\textbf{0.38}	&	4.05	&	-	&	1.46	&	2.83	\\
		\lightmidrule																									
		\rnn{}	&	2fc	&	8	&		&	-	&	2.55	&	16.48	&	-	&	38.75	&	111.25	&	-	&	44.47	&	113.77	\\
		&		&	16	&		&	-	&	2.10	&	3.60	&	-	&	15.21	&	64.85	&	-	&	12.89	&	57.02	\\
		&		&	32	&		&	-	&	1.75	&	1.99	&	-	&	5.98	&	23.85	&	-	&	12.10	&	12.28	\\
		&	\textbf{3fc}	&	8	&		&	-	&	-	&	-	&	-	&	-	&	-	&	-	&	-	&	-	\\
		&		&	16	&		&	-	&	\textbf{0.75}	&	2.50	&	-	&	9.97	&	56.01	&	-	&	5.28	&	45.47	\\
		&		&	\textbf{32}	&		&	-	&	1.06	&	0.76	&	-	&	\textbf{1.88}	&	15.88	&	-	&	5.95	&	\textbf{4.40}	\\
		\lightmidrule																									
		\lstm{}	&	2fc	&	8	&		&	-	&	0.83	&	2.34	&	-	&	12.32	&	44.67	&	-	&	16.23	&	32.87	\\
		&		&	16	&		&	-	&	0.60	&	0.77	&	-	&	3.85	&	17.51	&	-	&	3.87	&	4.98	\\
		&		&	32	&		&	-	&	0.57	&	0.94	&	-	&	3.55	&	9.21	&	-	&	3.09	&	6.78	\\
		&	\textbf{3fc}	&	8	&		&	-	&	-	&	-	&	-	&	7.23	&	-	&	-	&	-	&	-	\\
		&		&	16	&		&	-	&	0.44	&	0.50	&	-	&	2.05	&	15.69	&	-	&	\textbf{2.25}	&	3.22	\\
		&		&	\textbf{32}	&		&	-	&	\textbf{0.31}	&	0.57	&	-	&	\textbf{0.74}	&	5.10	&	-	&	2.27	&	2.98	\\
		\lightmidrule																									
		\gru{}	&	2fc	&	8	&		&	-	&	0.71	&	1.43	&	-	&	11.22	&	26.02	&	-	&	12.66	&	20.74	\\
		&		&	16	&		&	-	&	0.61	&	0.74	&	-	&	4.46	&	13.94	&	-	&	3.23	&	4.37	\\
		&		&	32	&		&	-	&	0.52	&	0.60	&	-	&	2.26	&	8.75	&	-	&	2.82	&	3.19	\\
		&	\textbf{3fc}	&	8	&		&	-	&	-	&	-	&	50.76	&	-	&	-	&	-	&	-	&	-	\\
		&		&	16	&		&	-	&	0.48	&	0.53	&	-	&	2.11	&	7.46	&	-	&	\textbf{2.49}	&	3.19	\\
		&		&	\textbf{32}	&		&	-	&	\textbf{0.38}	&	0.54	&	-	&	\textbf{0.65}	&	4.25	&	-	&	2.57	&	2.67	\\
		\lightmidrule																									
		\smallst{}	&		&	8	&		&	4.29	&	3.43	&	4.95	&	33.80	&	27.07	&	50.27	&	25.24	&	20.54	&	25.74	\\
		&		&	16	&		&	2.08	&	2.69	&	3.79	&	18.62	&	16.62	&	27.55	&	21.58	&	16.99	&	19.68	\\
		&		&	\textbf{32}	&		&	2.89	&	\textbf{1.91}	&	2.73	&	28.89	&	\textbf{8.40}	&	22.12	&	16.07	&	\textbf{11.98}	&	14.34	\\
		\lightmidrule																									
		\st{}	&		&	8	&		&	2.68	&	2.42	&	3.08	&	15.45	&	17.19	&	35.02	&	16.44	&	\textbf{18.22}	&	19.41	\\
		&		&	16	&		&	1.85	&	1.99	&	2.54	&	17.09	&	8.00	&	27.06	&	13.53	&	12.69	&	13.41	\\
		&		&	\textbf{32}	&		&	2.20	&	\textbf{1.79}	&	2.12	&	25.81	&	\textbf{2.38}	&	13.51	&	13.19	&	12.40	&	13.26	\\
	\end{tabular}
	\caption{Hyperparameter search results. We report mean squared validation error for all architectures, used encoder/decoder sub-networks (E/D), number of hidden dimensions (ND), and learning rate (LR) for three datasets. For \smallst{} and \st{}, we use the architecture used in prior work \cite{Lee2019} and, hence, do not use different encoder/decoder sub-networks. Cells without result indicate that the network did not converge to a reasonable result in this setup. We highlight the best result in each box.}
	\label{tab:hyperparameter_optimization}
\end{table*}

\clearpage

\begin{figure}[h!]
	\centering
	\begin{subfigure}{\linewidth}
		\begin{tikzpicture}
			\begin{axis}[
				width=\linewidth,
				xlabel=Set Utility,
				ylabel=Frequency]
				\addplot[ybar,bar width=0.1cm,fill=black] coordinates {(0 ,0) (1 ,0) (2 ,0) (3 ,1) (4 ,0) (5 ,1) (6 ,12) (7 ,10) (8 ,22) (9 ,27) (10,101) (11,121) (12,208) (13,234) (14,348) (15,617) (16,748) (17,1009) (18,1317) (19,1746) (20,2047) (21,2832) (22,3042) (23,3818) (24,4134) (25,4734) (26,4932) (27,5850) (28,5919) (29,6006) (30,6321) (31,5937) (32,5711) (33,5494) (34,5009) (35,4402) (36,4324) (37,3154) (38,2670) (39,2214) (40,1702) (41,1096) (42,1045) (43,468) (44,469) (45,148) (46,0) (47,0) (48,0) (49,0) (50,0)};
			\end{axis}
		\end{tikzpicture}
	\end{subfigure}
	
	\begin{subfigure}{\linewidth}
		\begin{tikzpicture}
			\begin{axis}[
				width=\linewidth,
				xlabel=Set Utility,
				ylabel=Frequency]
				\addplot[ybar,bar width=0.02cm,fill=black] coordinates {(0  ,0) (1  ,0) (2  ,0) (3  ,0) (4  ,0) (5  ,0) (6  ,0) (7  ,0) (8  ,0) (9  ,0) (10 ,0) (11 ,0) (12 ,0) (13 ,0) (14 ,0) (15 ,3) (16 ,8) (17 ,7) (18 ,4) (19 ,7) (20 ,16) (21 ,23) (22 ,24) (23 ,38) (24 ,56) (25 ,87) (26 ,94) (27 ,103) (28 ,138) (29 ,199) (30 ,262) (31 ,296) (32 ,354) (33 ,473) (34 ,532) (35 ,594) (36 ,699) (37 ,727) (38 ,867) (39 ,963) (40 ,1075) (41 ,1236) (42 ,1328) (43 ,1264) (44 ,1394) (45 ,1610) (46 ,1660) (47 ,1730) (48 ,1904) (49 ,1890) (50 ,2096) (51 ,2153) (52 ,2108) (53 ,2083) (54 ,2311) (55 ,2213) (56 ,2298) (57 ,2315) (58 ,2251) (59 ,2185) (60 ,2314) (61 ,2144) (62 ,2227) (63 ,2234) (64 ,2007) (65 ,1972) (66 ,2136) (67 ,1922) (68 ,1881) (69 ,1873) (70 ,1825) (71 ,1691) (72 ,1738) (73 ,1575) (74 ,1474) (75 ,1516) (76 ,1381) (77 ,1315) (78 ,1332) (79 ,1241) (80 ,1128) (81 ,1127) (82 ,963) (83 ,985) (84 ,1002) (85 ,960) (86 ,841) (87 ,921) (88 ,765) (89 ,673) (90 ,715) (91 ,615) (92 ,587) (93 ,628) (94 ,555) (95 ,539) (96 ,464) (97 ,420) (98 ,376) (99 ,403) (100,342) (101,339) (102,344) (103,279) (104,258) (105,253) (106,223) (107,185) (108,239) (109,192) (110,191) (111,211) (112,166) (113,160) (114,173) (115,145) (116,155) (117,145) (118,123) (119,115) (120,113) (121,87) (122,94) (123,108) (124,74) (125,87) (126,76) (127,64) (128,62) (129,61) (130,52) (131,41) (132,60) (133,33) (134,26) (135,22) (136,26) (137,29) (138,26) (139,25) (140,22) (141,27) (142,17) (143,17) (144,27) (145,20) (146,19) (147,18) (148,21) (149,16) (150,10) (151,11) (152,11) (153,16) (154,13) (155,11) (156,9) (157,11) (158,10) (159,8) (160,7) (161,13) (162,7) (163,9) (164,9) (165,9) (166,5) (167,4) (168,14) (169,2) (170,3) (171,4) (172,3) (173,1) (174,0) (175,4) (176,0) (177,3) (178,2) (179,2) (180,2) (181,1) (182,1) (183,4) (184,2) (185,5) (186,2) (187,0) (188,1) (189,3) (190,0) (191,0) (192,0) (193,0) (194,0) (195,1) (196,0) (197,0) (198,3) (199,0)};
			\end{axis}
		\end{tikzpicture}
		\label{fig:set_utility_distribution_syn}
	\end{subfigure}
	
	\begin{subfigure}{\linewidth}
		\begin{tikzpicture}
			\begin{axis}[
				width=\linewidth,
				xlabel=Set Utility,
				ylabel=Frequency]
				\addplot[ybar,bar width=0.05cm,fill=black] coordinates {(0  ,0) (1  ,0) (2  ,0) (3  ,0) (4  ,0) (5  ,1) (6  ,2) (7  ,10) (8  ,10) (9  ,14) (10 ,24) (11 ,25) (12 ,69) (13 ,32) (14 ,99) (15 ,140) (16 ,49) (17 ,183) (18 ,168) (19 ,107) (20 ,182) (21 ,271) (22 ,183) (23 ,341) (24 ,411) (25 ,462) (26 ,629) (27 ,460) (28 ,1196) (29 ,887) (30 ,1100) (31 ,1698) (32 ,1287) (33 ,1467) (34 ,1618) (35 ,1592) (36 ,1598) (37 ,1910) (38 ,1588) (39 ,2119) (40 ,2008) (41 ,1754) (42 ,2471) (43 ,2351) (44 ,2872) (45 ,3167) (46 ,2604) (47 ,3993) (48 ,2754) (49 ,3604) (50 ,3202) (51 ,3380) (52 ,2756) (53 ,3575) (54 ,2159) (55 ,2520) (56 ,2916) (57 ,1402) (58 ,2595) (59 ,1951) (60 ,2366) (61 ,1805) (62 ,2108) (63 ,1795) (64 ,1831) (65 ,1938) (66 ,2113) (67 ,1757) (68 ,1458) (69 ,914) (70 ,893) (71 ,585) (72 ,958) (73 ,304) (74 ,932) (75 ,0) (76 ,168) (77 ,159) (78 ,181) (79 ,456) (80 ,179) (81 ,459) (82 ,148) (83 ,149) (84 ,178) (85 ,172) (86 ,0) (87 ,0) (88 ,0) (89 ,0) (90 ,0) (91 ,0) (92 ,0) (93 ,0) (94 ,0) (95 ,28) (96 ,0) (97 ,0) (98 ,0) (99 ,0) (100,0)};
			\end{axis}
		\end{tikzpicture}
	\end{subfigure}
	
	\caption{Distribution of set utilities for different datasets. Top: \mnistredshort{}, middle: \mnistsynshort{}, bottom: \mnistredsynshort{}}
	\label{fig:set_utility_distributions}
\end{figure}


\section{Results without Enforcing Non-negative Intermediate Results}
In Table~\ref{tab:training_results_without_abs}, we report results without enforcing non-negative intermediate results by removing $\abs{.}$ in Equation~8. We observe that the accuracy can further improve in several cases and that we achieve even new best results in three datasets (\mnistredshort{}, \mnistredsynshort{}, \fmnisttrishort{}). These results indicate that removing the monotonicity constraint can be helpful to achieve better results. However, it should be considered Choquet capacities always enforce monotonicity. Hence, by removing the monotonicity constraint the proposed idea becomes less strongly related to Choquet capacities.

\begin{table}[H]
	\centering
	\begin{tabular}{l r r r r}
		& \multicolumn{1}{c}{\mnistredshort{}} &  \multicolumn{1}{c}{\mnistsynshort{}} & \multicolumn{1}{c}{\mnistredsynshort{}} & \multicolumn{1}{c}{\fmnisttrishort{}} \\
		\toprule
		\rnnc{} (w/o abs.)	&	0.40	&	0.79	&	2.12	&	\textbf{0.98} \\
		\lstmc{} (w/o abs.)	&	0.22	&	0.66	&	1.28	&	1.03 \\
		\gruc{} (w/o abs.)	&	\textbf{0.21}	&	0.75	&	\textbf{1.05}	&	1.02 \\
	\end{tabular}
	\caption{Results without absolute value computation.}
	\label{tab:training_results_without_abs}
\end{table}

\section{Additional Results for Varying Set Sizes}
To evaluate the sensitivity of the models to varying set sizes, we generated a dataset with $6,8,10,12,$ and $14$ instances. In addition to Table~\ref{tab:varying_set_sizes_results}, we report the median of the mean squared error for experiments with varying set sizes in Table~\ref{tab:varying_set_sizes_results_median}. The results confirm previously made observations that Capacity networks perform better than their non-Capacity counterparts in these experiments.

\begin{table}[h!]
	\centering
	\begin{tabular}{l @{\hspace{0.3cm}} r @{\hspace{0.3cm}} r @{\hspace{0.3cm}} r @{\hspace{0.3cm}} r}
		& \multicolumn{1}{c}{\mnistredshort{}} &  \multicolumn{1}{c}{\mnistsynshort{}} & \multicolumn{1}{c}{\mnistredsynshort{}} &  \multicolumn{1}{c}{\fmnisttrishort{}} \\
		\toprule
		\rnn	&	1.31	&	3.55	&	9.12	&	107.51	\\
		\lstm	&	0.42	&	2.14	&	2.76	&	2.16	\\
		\gru	&	0.43	&	1.01	&	2.45	&	105.26	\\
		\lightmidrule
		\rnnc	&	0.36	&	0.98	&	2.92	&	38.07	\\
		\lstmc	&	\textbf{0.25}	&	0.77	&	\textbf{1.56}	&	1.89	\\
		\gruc	&	0.29	&	\textbf{0.40}	&	1.75	&	\textbf{1.57}	\\
		\bottomrule
	\end{tabular}
	\caption{Mean squared errors for varying set sizes.}
	\label{tab:varying_set_sizes_results}
\end{table}

\begin{table}[h!]
	\centering
	\begin{tabular}{l r r r r}
		& \multicolumn{1}{c}{\mnistredshort{}} &  \multicolumn{1}{c}{\mnistsynshort{}} & \multicolumn{1}{c}{\mnistredsynshort{}} &  \multicolumn{1}{c}{\fmnisttrishort{}} \\
		\toprule
		
		\rnn	&	1.20	&	4.06	&	9.51	&	107.51	\\
		\lstm	&	0.40	&	1.59	&	2.77	&	2.11	\\
		\gru	&	0.40	&	0.98	&	2.56	&	105.26	\\
		\lightmidrule									
		\rnnc	&	0.39	&	1.03	&	3.01	&	3.44	\\
		\lstmc	&	\textbf{0.26}	&	0.69	&	\textbf{1.52}	&	1.82	\\
		\gruc	&	0.28	&	\textbf{0.40}	&	1.78	&	\textbf{1.49}	\\
		
		\bottomrule
	\end{tabular}
	\caption{Median for experiments with varying set sizes.}
	\label{tab:varying_set_sizes_results_median}
\end{table}

\begin{table*}[h!]
	\centering
	\begin{tabular}{l r r r r r r r r r}
		& \multicolumn{3}{c}{\mnistredshort{}} &  \multicolumn{3}{c}{\mnistsynshort{}} & \multicolumn{3}{c}{\mnistredsynshort{}} \\
		
		\cmidrule(lr){2-4} \cmidrule(lr){5-7} \cmidrule(lr){8-10}
		& 30k & 50k & 70k & 30k & 50k & 70k & 30k & 50k & 70k \\	
		\midrule
		\dset	&	1.54	&	9.84	&	0.35	&	10.85	&	6.20	&	4.24	&	111.38	&	6.03	&	55.68	\\
		\att	&	1.88	&	0.42	&	0.35	&	5.56	&	2.70	&	2.18	&	65.61	&	4.42	&	2.24	\\
		\smallst&	4.61	&	3.30	&	2.50	&	34.03	&	19.08	&	12.53	&	27.07	&	20.39	&	17.28	\\
		\st		&	3.50	&	2.62	&	1.95	&	20.81	&	13.47	&	9.36	&	18.63	&	17.17	&	14.97	\\
		\lightmidrule
		\rnn	&	1.42	&	1.01	&	1.04	&	11.68	&	5.91	&	3.02	&	63.87	&	8.27	&	7.88	\\
		\lstm	&	0.84	&	0.60	&	0.42	&	5.05	&	2.41	&	1.27	&	5.71	&	3.55	&	2.58	\\
		\gru	&	0.76	&	0.50	&	0.40	&	4.18	&	1.95	&	1.01	&	6.19	&	4.01	&	2.82	\\
		\lightmidrule
		\rnnc	&	0.98	&	0.55	&	0.50	&	5.47	&	2.14	&	0.91	&	9.89	&	4.60	&	3.07	\\
		\lstmc	&	\textbf{0.70}	&	\textbf{0.37}	&	\textbf{0.29}	&	4.33	&	1.60	&	0.82	&	\textbf{6.14}	&	\textbf{2.75}	&	\textbf{1.73}	\\
		\gruc	&	0.76	&	0.42	&	0.45	&	\textbf{3.73}	&	\textbf{1.37}	&	\textbf{0.66}	&	5.33	&	2.89	&	1.94	\\
		\bottomrule
	\end{tabular}
	\caption{Mean squared error of three architecture vs. each architecture equipped with our newly proposed inductive bias on three datasets with different non-additive utility functions for four different training set sizes.}
	\label{tab:app_var_training_data}
\end{table*}

\section{Additional Results for Experiments with Varying Amounts of Training Data}
\label{sec:app_var_training_data}
Table~\ref{tab:app_var_training_data} contains results for parallel architectures for the experiments with varying amounts of training data. In contrast to the main paper, in which we focused on the difference between Capacity networks and their non-Capacity counterparts by showing relative values, we new report absolute values.

\section{Quantitative Results for Figure~\ref{fig:permutation_sensivity}}
In Table~\ref{tab:permutation_sensivity_variance}, we report the quantitative results for the illustration in Figure~\ref{fig:permutation_sensivity}.

\begin{table}[h!]
	\centering
	\begin{tabular}{l @{\hspace{0.2cm}} c @{\hspace{0.2cm}} c @{\hspace{0.2cm}} c @{\hspace{0.2cm}} c @{\hspace{0.2cm}} c @{\hspace{0.2cm}} c}
		& {\small \rnn}  &  {\small \rnnc} & {\small \lstm} & {\small \lstmc} & {\small \gru} & {\small \gruc} \\
		\cmidrule(lr){2-3} \cmidrule(lr){4-5} \cmidrule(lr){6-7}
		Median&0.88	&\textbf{0.39}	&0.34	&\textbf{0.23}	&0.35	&\textbf{0.28}	\\
		Mean  &0.82	&\textbf{0.38}	&0.34	&\textbf{0.22}	&0.34	&\textbf{0.28}	\\
		StDev &0.13	&\textbf{0.06}	&\textbf{0.03}	&\textbf{0.03}	&0.05	&\textbf{0.04}	\\
	\end{tabular}
	\caption{Median, mean, and standard deviation (StDev) for the results in Figure~\ref{fig:permutation_sensivity}.}
	\label{tab:permutation_sensivity_variance}
\end{table}

\section{Additional Improved Interpretability Illustrations}
\label{sec:app_improved_interpretability}
Table~\ref{tab:further_interpretability_examples} provides more insights on learned intermediate results similar to Table~\ref{tab:interpretability}. Specifically, we provide two more examples for the \mnistsynshort{} and the \mnistredsynshort{} dataset. Similar to the results in Table~\ref{tab:interpretability}, it can be observed that Capacity networks learn reasonable intermediate results without any instance-level supervision.

\begin{table}[h!]
	\centering
	\begin{tabular}{r c c c c c r}
		Input &
		\scalebox{0.40}{\includegraphics{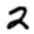}} &
		\scalebox{0.40}{\includegraphics{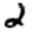}} &
		\scalebox{0.40}{\includegraphics{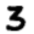}} & 
		\scalebox{0.40}{\includegraphics{images/mnist_6_1.png}} &
		\scalebox{0.40}{\includegraphics{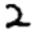}} & \multicolumn{1}{c}{$\sum$} \\
		
		Expected & 2.00    & 4.00    & 3.00    & 6.00    & 6.00   & 21.00  \\
		\midrule
		\multicolumn{1}{c}{\rnnc}    & 1.99 & 4.00 & 2.98 & 5.95 & 5.65 & 20.57  \\
		\multicolumn{1}{c}{$\Delta$} & 0.01 & 0.00 & 0.02 & 0.05 & 0.35 & 0.43 \\
		\midrule
		\multicolumn{1}{c}{\lstmc}   & 2.02 & 3.97 & 3.04 & 6.00 & 6.21 & 21.24 \\
		\multicolumn{1}{c}{$\Delta$} & 0.02 & 0.03 & 0.04 & 0.00 & 0.21 & 0.30 \\
		\midrule
		\multicolumn{1}{c}{\gruc}    & 2.03 & 3.97 & 2.92 & 5.95 & 6.38 & 21.25 \\
		\multicolumn{1}{c}{$\Delta$} & 0.03 & 0.03 & 0.08 & 0.05 & 0.38 & 0.57 \\
		
		\multicolumn{7}{c}{ } \\
		Input &
		\scalebox{0.40}{\includegraphics{images/mnist_7_1.png}} &
		\scalebox{0.40}{\includegraphics{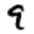}} &
		\scalebox{0.40}{\includegraphics{images/mnist_2_1.png}} & 
		\scalebox{0.40}{\includegraphics{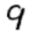}} &
		\scalebox{0.40}{\includegraphics{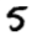}} & \multicolumn{1}{c}{$\sum$} \\
		
		Expected & 7.00    & 9.00    & 12.00    & 0.00    & 5.00 & 33.00   \\
		\midrule
		\multicolumn{1}{c}{\rnnc}    & 6.86 & 8.97 & 11.92 & 0.11 & 5.03 & 32.89 \\
		\multicolumn{1}{c}{$\Delta$} & 0.14 & 0.03 & 0.08 & 0.11 & 0.03 & 0.39 \\
		\midrule
		\multicolumn{1}{c}{\lstmc}   & 6.09 & 8.50 & 12.27 & 0.57 & 5.70 & 33.13 \\
		\multicolumn{1}{c}{$\Delta$} & 0.91 & 0.50 & 0.27 & 0.57 & 0.70 & 2.95 \\
		\midrule
		\multicolumn{1}{c}{\gruc}    & 6.81 & 8.59 & 11.91 & 0.24 & 5.01 & 32.56 \\
		\multicolumn{1}{c}{$\Delta$} & 0.19 & 0.41 & 0.09 & 0.24 & 0.01 & 0.94 \\
	\end{tabular}
	\caption{Illustration of improved interpretability in the \mnistsynshort{} (top) and \mnistredsynshort{} (bottom) dataset. We show the MNIST input instances (first row), the expected intermediate results (second row), the predicted intermediate results, and the difference to the expected value ($\Delta$ rows).}
	\label{tab:further_interpretability_examples}
\end{table}

\begin{table*}
	\centering
	\begin{tabular}{l r r r r r r r r r}
		& \multicolumn{3}{c}{\mnistredshort{}} &  \multicolumn{3}{c}{\mnistsynshort{}} & \multicolumn{3}{c}{\mnistredsynshort{}} \\
		
		\cmidrule(lr){2-4} \cmidrule(lr){5-7} \cmidrule(lr){8-10}
		& 30k & 50k & 70k & 30k & 50k & 70k & 30k & 50k & 70k \\	
		\cmidrule(lr){2-10}
		
		\rnn{} & 60.61 &	68.93 &	73.22 	&	17.01 &	29.06 &	40.13 	&	\textbf{46.70} &	54.52 &	53.71 \\
		\rnnc{}	& \textbf{71.42} &	\textbf{92.31} &	\textbf{96.13} 	&	\textbf{33.04} &	\textbf{71.37} &	\textbf{74.30} 	&	30.67 &	\textbf{58.74} &	\textbf{76.51}
		
	\end{tabular}
	\caption{Additional accuracy results for Table~\ref{tab:var_training_data}.}
	\label{tab:var_training_data_accuracy}
\end{table*}

\section{Evaluating Intermediate Results of Non-Capacity Networks}
In principle, it is possible to obtain results similar to the intermediate results generated by Capacity networks by applying non-Capacity networks to subsets of the input. To this end, we train the networks as described previously an apply the decoder of the networks to the subsets $\set{x_1}, \set{x_1, x_2}, ...$. The added valued of $\instance_i$ can then be computed according to $\decoder(\set{x_1, \dots, x_i}) - \decoder(\set{x_1, \dots, x_{i-1}})$. However, unlike Capacity networks, the decoder of non-Capacity networks has never been trained to generate good intermediate results during training. Hence, it is unlikely that this strategy performs well. To demonstrate this, we apply the idea to the \mnistsynshort{} and \fmnisttrishort{} datasets and compute the mean absolute error of the obtained intermediate results with respect to the expected intermediate results. Table~\ref{tab:intermediate_results_eval_mnistredsyn} shows that non-Capacity indeed generate very poor intermediate results, specifically in the \mnistsynshort{} dataset, compared to Capacity networks.

\begin{table}
	\centering
	\begin{tabular}{l @{\hspace{0.2cm}} c @{\hspace{0.2cm}} c @{\hspace{0.2cm}} c @{\hspace{0.2cm}} c @{\hspace{0.2cm}} c @{\hspace{0.2cm}} c}
		& \rnn  &  \rnnc & \lstm & \lstmc & \gru & \gruc \\
		\cmidrule(lr){2-3} \cmidrule(lr){4-5} \cmidrule(lr){6-7}
		\mnistsynshort{} & 12.45	&	\textbf{0.12}	&	17.56	&	\textbf{0.11}	&	13.25	&	\textbf{0.09}\\
		\fmnisttrishort{} & 3.84	&	\textbf{1.10}	&	3.47	&	\textbf{0.81}	&	11.75	&	\textbf{1.15}\\
	\end{tabular}
	\caption{Mean absolute error of intermediate results.}
	\label{tab:intermediate_results_eval_mnistredsyn}
\end{table}

\section{Accuracy Results for Table~\ref{tab:var_training_data}}
In Table~\ref{tab:var_training_data_accuracy}, we report the accuracy results for the \rnn{} and the \rnnc{} architecture. The results show that not only the mean squared error improves substantially, but also the achieved accuracy if the task is understood as classification problem. It is likely that better results can be achieved if a classification loss such as the cross-entropy loss are used if the accuracy should be maximized.

\end{document}